\pdfoutput=1
\documentclass[11pt]{article}

\usepackage[final]{acl}
\usepackage{multirow}
\usepackage{booktabs}
\usepackage{array}
\usepackage{adjustbox} 
\usepackage{amsmath}

\usepackage{times}
\usepackage{latexsym}

\usepackage[T1]{fontenc}
\usepackage[utf8]{inputenc}

\usepackage{microtype}

\usepackage{inconsolata}

\usepackage{graphicx}

\usepackage{makecell}
\newcolumntype{x}[1]{>{\centering\arraybackslash}m{#1}}

\definecolor{thao}{HTML}{F59CA1}

\definecolor{Tiara}{HTML}{FFA500}

\definecolor{Chang}{HTML}{0000FF}

\usepackage{xparse}
\usepackage{authblk}

\usepackage{xcolor}
\definecolor{red}{rgb}{0.99, 0.02, 0.02}
\NewDocumentCommand{\heng}
{ mO{} }{\textcolor{red}{\textsuperscript{\textit{Heng}}\textsf{\textbf{\small[#1]}}}}

\NewDocumentCommand{\carl}
{ mO{} }{\textcolor{blue}{\textsuperscript{\textit{Carl}}\textsf{\textbf{\small[#1]}}}}

\newcommand{\model}{\textbf{GLaD}}

\title{GLaD: Synergizing Molecular Graphs and Language Descriptors for Enhanced Power Conversion Efficiency Prediction \\in Organic Photovoltaic Devices}

\author[1]{Thao Nguyen}
\author[2]{Tiara Torres-Flores}
\author[2]{Changhyun Hwang}
\author[1]{Carl Edwards}
\author[2]{Ying Diao}
\author[1]{Heng Ji}

\affil[1]{Siebel School of Computing and Data Science, University of Illinois Urbana-Champaign}
\affil[2]{Department of Chemical \& Biomolecular Engineering, University of Illinois Urbana-Champaign}

\affil[ ]{\texttt{\{thaotn2, tiaract2, chwang12, cne2, yingdiao, hengji\}@illinois.edu}}

\begin{document}
\maketitle
\begin{abstract}
This paper presents a novel approach for predicting Power Conversion Efficiency (PCE) of Organic Photovoltaic (OPV) devices, called \model{}: synergizing molecular \underline{G}raphs and \underline{La}nguage \underline{D}escriptors for enhanced PCE prediction. 
Due to the lack of high-quality experimental data, we collect a dataset consisting of 500 pairs of OPV donor and acceptor molecules along with their corresponding PCE values, which we utilize as the training data for our predictive model.
In this low-data regime, \model{} leverages properties learned from large language models (LLMs) pretrained on extensive scientific literature to enrich molecular structural representations, allowing for a multimodal representation of molecules.
\model{} achieves precise predictions of PCE, thereby facilitating the synthesis of new OPV molecules with improved efficiency.
Furthermore, \model{} showcases versatility, as it applies to a range of molecular property prediction tasks (BBBP, BACE, ClinTox and SIDER~\cite{wu2018moleculenet}), not limited to those concerning OPV materials.
Especially, \model{} proves valuable for tasks in low-data regimes within the chemical space, as it enriches molecular representations by incorporating molecular property descriptions learned from large-scale pretraining. This capability is significant in real-world scientific endeavors like drug and material discovery, where access to comprehensive data is crucial for informed decision-making and efficient exploration of the chemical space.
% Furthermore, our framework showcases versatility, indicating its broader applicability to a range of molecular property prediction tasks, not limited to those concerning OPV materials.

% \carl{Focus on how methodology enables us to leverage ``learned properties and associations from large-scale pretraining on the scientific literature via a large language model.'' Also talk about how our method is especially valuable on low-data regime tasks within the chemical space by leveraging large-scale pretraining. These tasks are abundant within real-world scientific problems such as drug discovery (cite) or ???. }

\end{abstract}

\begin{figure}
    \centering
    \includegraphics[width=0.49\textwidth]{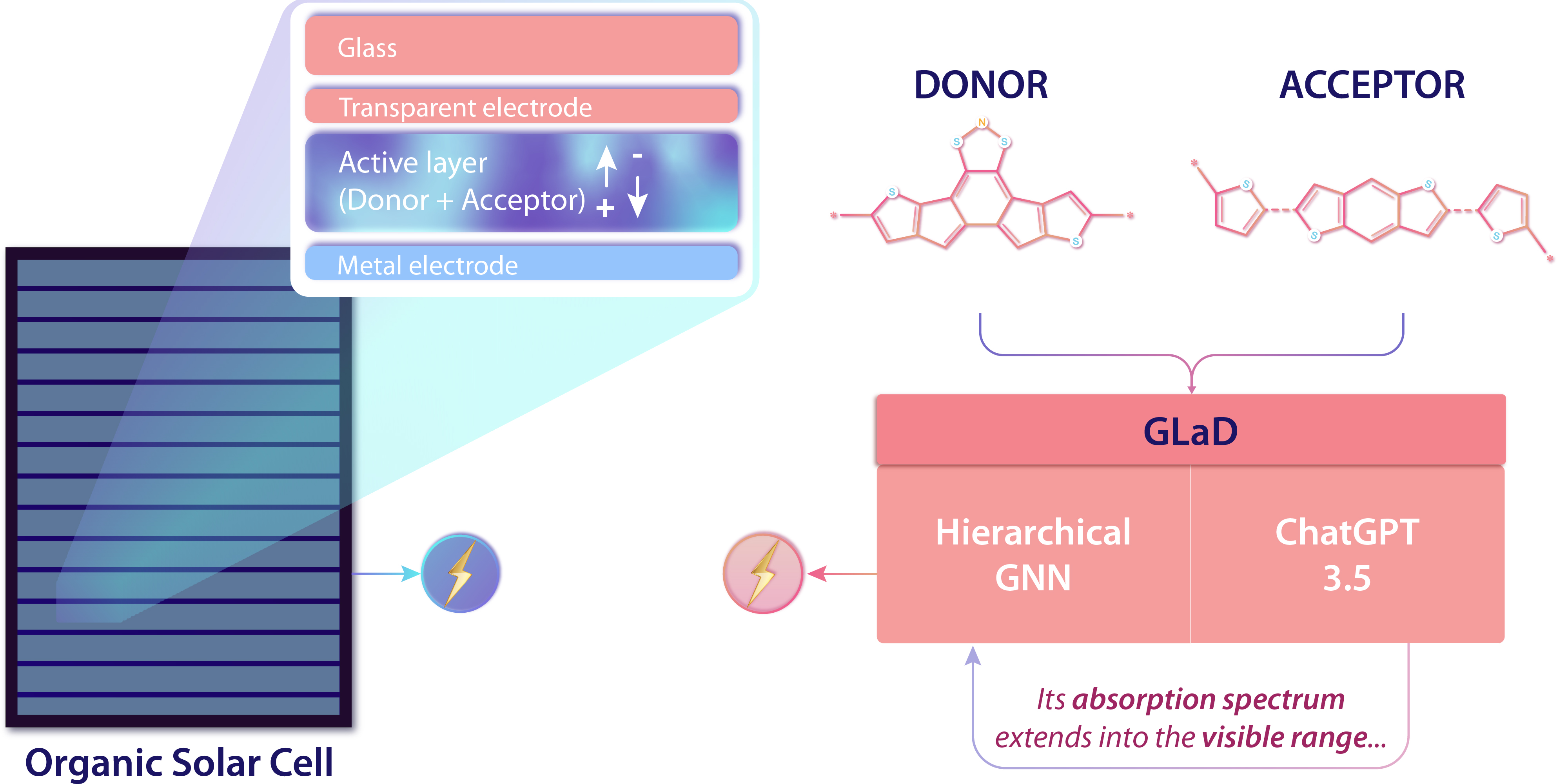}
    \caption{Overview of the \model{} PCE prediction framework.}
    \label{fig:overview}
    \vspace{-1em}
\end{figure}

\section{Introduction}
In materials science, the design of novel materials for organic solar cells (OSCs) is a vibrant area of research, as OSCs offer advantages such as being low-cost, flexible, and lightweight; unfortunately, they also suffer from drawbacks such as limited lifespan and poor stability~\cite{abdulrazzaq2013organic, kumar2012recent, burlingame2020stability}. Addressing these drawbacks necessitates the optimization of materials for OSCs, which requires quick and accurate prediction of Power Conversion Efficiency (PCE) in OSC devices to assess the quality of new candidates.
% \Tiara{I do not think we are addressing the quickness or accuracy of PCE measurement in our work. I would change this sentence to: Addressing these challenges requires the optimization of materials development for OSCs, where the power conversion efficiency (PCE) is crucial for assessing the quality of new candidates.}
% \thao{Our PCE prediction method is quicker than using Scharber's model, and demonstrated to be accurate by experiments}

Various machine learning algorithms have been used to predict PCE of OPV devices using different datasets. Notably, the Harvard Clean Energy Project Database (CEPDB)~\cite{hachmann2011harvard} and the Harvard Organic Photovoltaic Dataset (HOPV)~\cite{lopez2016harvard} are among the most significant public datasets in this domain. Previous studies have primarily utilized CEPDB, which comprises 2.3 million donor molecules and their corresponding PCE values calculated using Scharber's model~\citep{scharber2006design}.
While training with computationally derived PCEs offers the advantage of large, standardized datasets with controlled parameters, these values often poorly correlate with experimental measurements, diminishing their practicality~\cite{greenstein2022computational}.
The HOPV dataset contains experimental PCE data for 350 different OPV donors that have been collected from various studies in the literature by Lopez et al~\cite{lopez2016harvard}, yet it lacks data on newer OPV molecules introduced after 2015, a period during which significant advancements in OPV technology achieved PCE values of up to 20\%~\cite{guan2024opv20,fu2023opv19,elamine2023opv20,cui2021opv19}. Therefore, to expedite the development of cutting-edge OPV materials, this study concentrates on predicting the experimental PCE values of recently developed OPV devices only.

\begin{table}[!t]
    \centering
    \footnotesize
    \setlength{\tabcolsep}{5pt}
    \resizebox{0.9\columnwidth}{!}{
    \begin{tabular}{c l x{0.04\textwidth} x{0.04\textwidth} x{0.04\textwidth} x{0.04\textwidth}}
        \toprule
        \multicolumn{2}{l}{} & \textbf{All} & \textbf{Train} & \textbf{Val} & \textbf{Test} \\
        \multicolumn{2}{l}{\textbf{\#samples (D-A pairs)}} & \textbf{500} & \textbf{400} & \textbf{50} & \textbf{50} \\
        \midrule 
        \multirow{3}{*}{\textbf{\#molecules}} & All & 403 & 338 & 61 & 58 \\
        & Donor & 203 & 165 & 36 & 32 \\
        & Acceptor & 252 & 212 & 31 & 34 \\
        \cmidrule{2-6} 
        \multirow{3}{*}{\textbf{\makecell{\#functional \\ modules}}} & All & 250 & 231 & 68 & 68 \\
        & Donor & 149 & 136 & 38 & 38 \\
        & Acceptor & 192 & 174 & 43 & 43 \\
        \cmidrule{2-6} 
        \multirow{3}{*}{\textbf{\makecell{Tanimoto \\ distance}}} & All & 0.67 & 0.67 & 0.69 & 0.67 \\
        & Donor & 0.65 & 0.64 & 0.65 & 0.69 \\
        & Acceptor & 0.59 & 0.59 & 0.64 & 0.63 \\
        \midrule
        \textbf{PCE range} & & [2.5, 19.6] & [2.5, 19.6] & [6.0, 18.69] & [5.1, 17.1] \\
        \bottomrule
        
    \end{tabular}}
    \caption{Statistics of our collected dataset}
    \label{table:data_stats}
\end{table}

In this work, we present a novel approach, named \model{}, for accurately predicting PCE of OPV devices based on pairs of donor and acceptor molecules. To achieve this, we collected a dataset comprising 500 pairs of donors and acceptors from the literature for training our models.

\model{} addresses a key challenge in predicting PCE of OPV devices: the need for a comprehensive understanding of molecular function-structure relationships. To tackle this, chemists typically focus on the functional modules of a molecule and rely on supplementary sources like textbooks for a more comprehensive understanding of the molecule's properties. Inspired by these insights, we decompose molecules into their functional modules and integrate structural descriptors extracted by a Graph Neural Network (GNN) with textual descriptions generated by LLMs trained on extensive scientific literature. This approach aims to provide a comprehensive representation of the functional modules.
% Our work is the first attempt to harness the vast knowledge provided by LLMs about the properties of functional groups to create a comprehensive multimodal representation of molecules, ultimately improving the accuracy of PCE predictions for OPV devices. 
After acquiring the structural and textual descriptors of the those modules, we fuse them to form a multimodal representation. Subsequently, representations of functional modules are fed into a molecule-level GNN model to predict PCE.
Figure~\ref{fig:overview} illustrates the overview of \model{} in the PCE prediction task.

We assessed the performance of \model{} using our collected dataset, HOPV and the MoleculeNet benchmark~\cite{wu2018moleculenet}. Our results demonstrate that \model{} accurately predicts PCE values for OPV devices.
Notably, incorporating textual descriptors alongside structural descriptors enhances the model's performance, with the coefficient of determination (R$^2$) score increasing by 0.103 ($\pm$ 0.04) in our collected dataset.
For HOPV dataset, we obtain an R$^2$ score improvement of 0.135 compared to the baseline~\cite{eibeck2021predicting}, showcasing state-of-the-art performance on this dataset.
Furthermore, \model{} exhibits high accuracy in predicting molecular properties across various tasks (such as BBBP, BACE, ClinTox and SIDER~\cite{wu2018moleculenet}), suggesting its applicability beyond OPV-related tasks. \\% \model{} is particularly valuable for training models in low-data regime.
% In summary, our contribution comprises four key aspects:
Our contributions are summarized as follows:
\begin{itemize}
    \item We curate an up-to-date dataset comprising 500 pairs of donor and acceptor molecules for PCE prediction task.
    \item We develop a novel method, \model{}, that leverages learned knowledge from pretrained LLMs to generate textual descriptions for functional modules (molecular fragments) and integrates them with structural descriptors to enrich molecule representation. This approach accurately predicts PCE, achieving high R$^2$ scores in both our dataset and the HOPV dataset.
    \item \model{} is the first model to use a hierarchical GNN approach to integrate textual descriptions of molecular fragments (functional modules) rather than entire molecule descriptions. This improves the robustness and flexibility of our approach on unknown molecules. We conducted a study of language model-generated textual descriptions and found 88\% of them to be accurate when evaluated by PhD-level domain experts.  
    % \item By incorporating textual descriptors, we demonstrate a significant improvement in predictive performance.%, enhancing the model's R$^2$ score by 0.103. 
    \item Our method exhibits promising results in other molecular property prediction tasks, indicating its broad applicability beyond PCE prediction.

\end{itemize}

\section{Background}
OSCs represent a promising class of emerging photovoltaic technologies that generate electricity from sunlight using multiple layers. In these cells, excitons (hole-electron pairs) are produced at the interface of the active layer, typically composed of a donor-acceptor (D-A) material blend of carbon-based molecules or polymers. The donor material absorbs light and generates excitons, while the acceptor material facilitates their dissociation into free charge carriers. D-A molecules thus play a crucial role in determining PCE of OPV devices by influencing light absorption, exciton dissociation, charge transport, and active layer morphology~\citep{zhang2021DAmaterials}.  Optimizing molecular combinations can achieve broad absorption spectra, efficient exciton separation, balanced charge mobility, and ideal nanoscale phase separation, all of which enhance PCE.

Traditional methods for designing D-A combinations focus on a limited range of chemistries and typically require expert knowledge, restricting the potential for high-throughput screening~\citep{hachmann2011harvard}. However, computational modeling and machine learning offer a powerful alternative by accurately predicting the PCE of new D-A materials through the analysis of their molecular properties. This accelerates the discovery of high-performance OPV materials and broadens the range of structures explored for improved efficiency and stability.

\section{Related Work}
\subsection{PCE Prediction}

\begin{figure}[!t]
    \centering
    \includegraphics[width=\columnwidth]{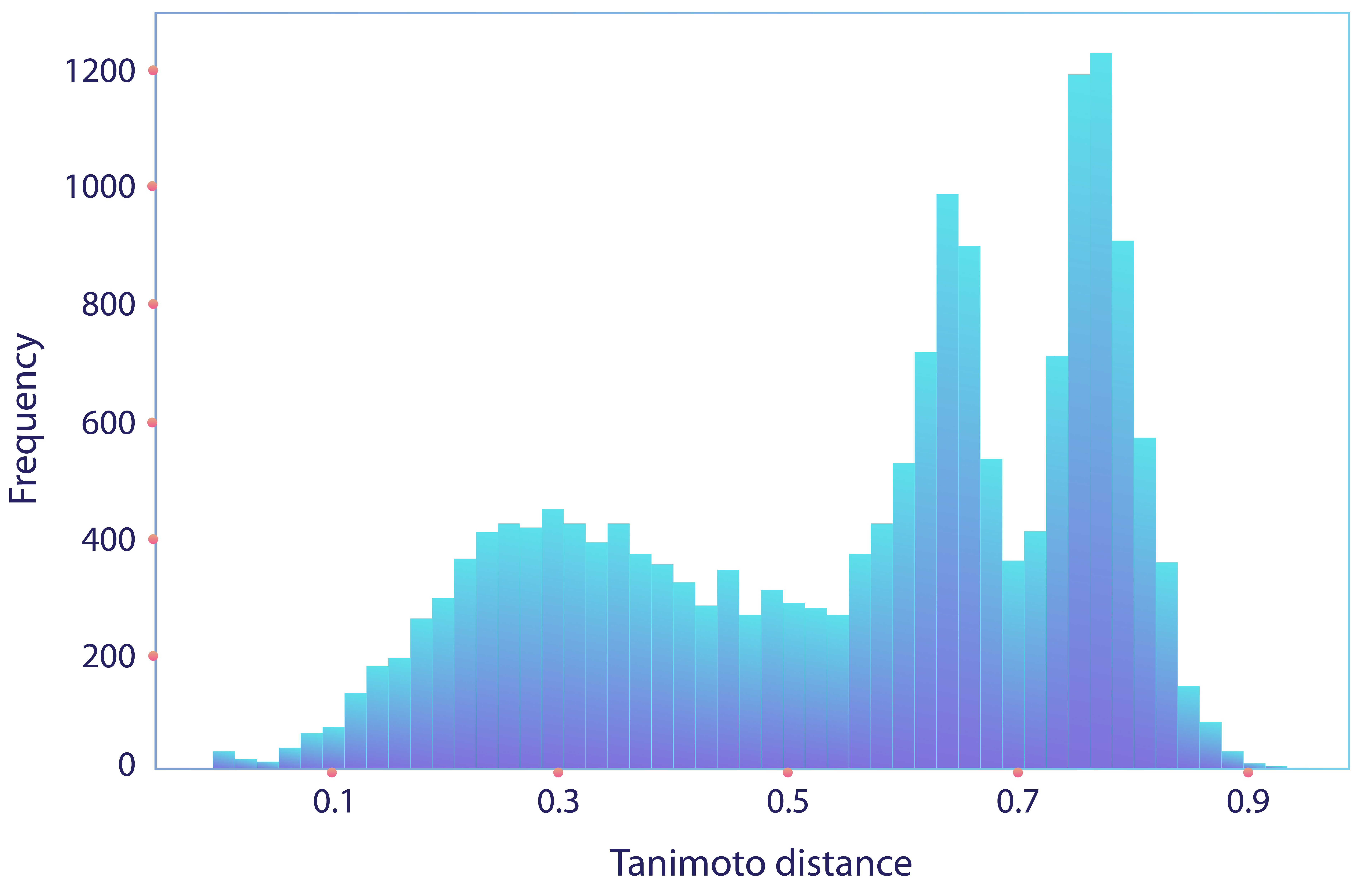}
    \caption{Distribution of pairwise Tanimoto distances among molecules in the collected dataset.}
    \label{fig:tanimoto}
\end{figure}

\begin{figure}[!t]
    \centering
    \includegraphics[width=\columnwidth]{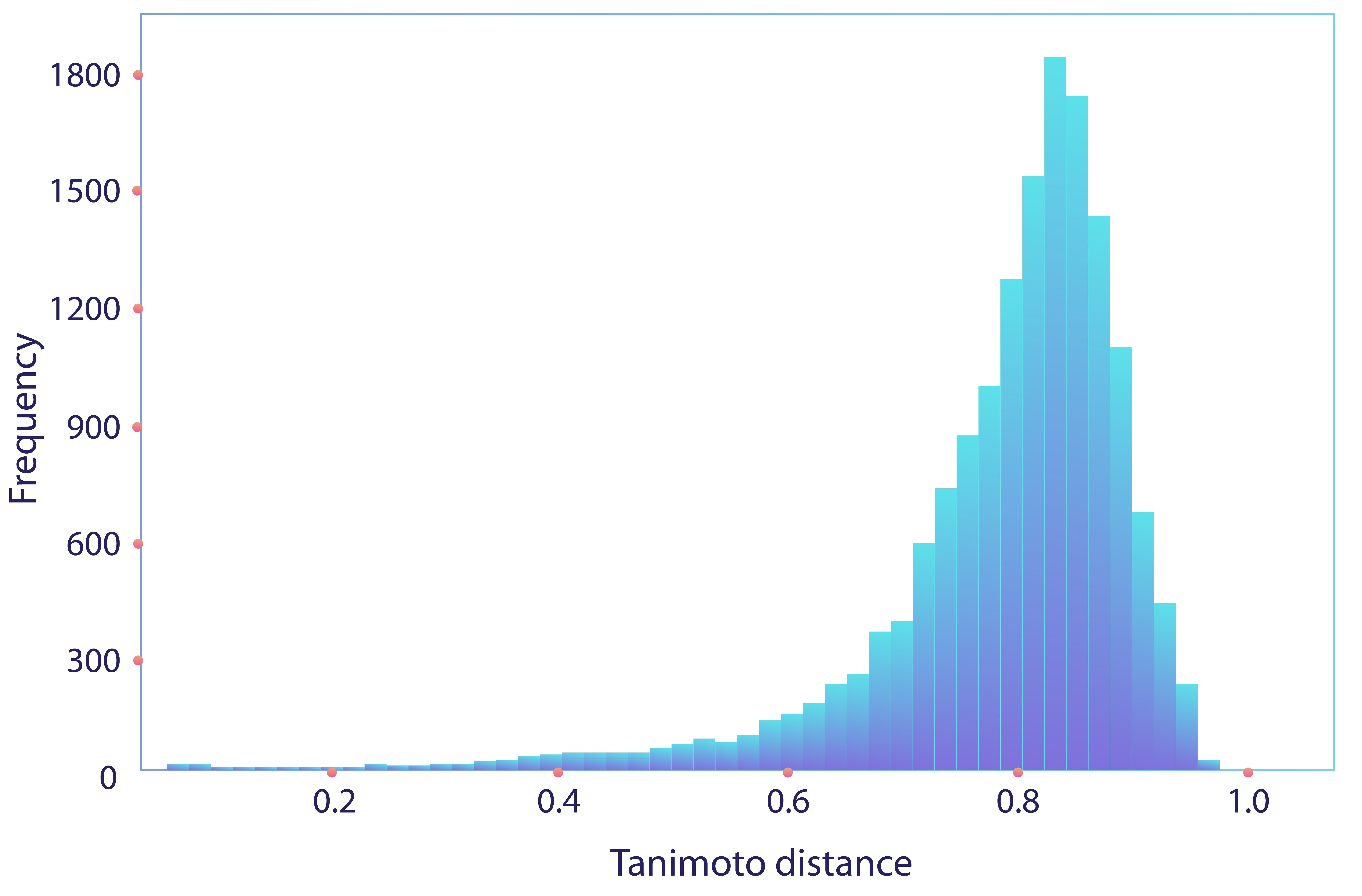}
    \caption{Distribution of pairwise Tanimoto distances among molecules in the HOPV dataset.}
    \label{fig:hopv_tanimoto}
\end{figure}

The prediction of PCE has drawn considerable interest, particularly with the emergence of large datasets like CEPDB~\cite{hachmann2011harvard} and the experimental dataset HOPV~\cite{lopez2016harvard}.
Various methods are employed for this task, including quantum chemical calculations and machine learning (ML) techniques.
Quantum chemical methods estimate PCE using Scharber's model~\cite{scharber2006design}. This model predicts PCE of a specific OPV design based on parameters calculated by density functional theory (DFT). However, DFT calculations require significant computational time, which makes them unsuitable for quick screening~\cite{adamo2013calculations}, and there is a discrepancy between the predictions of Scharber's model and actual experimental results~\cite{greenstein2022computational}.
% However, this method requires significant computational time and there is a discrepancy between the predictions of Scharber's model and actual experimental results~\cite{greenstein2022computational}. 
On the other hand, ML techniques are commonly used to explore the relationships between OPV performance and material properties more quickly and accurately~\cite{greenstein2023screening, wu2020machine}.\\
% Therefore, ML presents a practical option for efficient and cost-effective screening of OPV materials, as well as broader applications in material and drug design.
Many studies have focused on predicting PCE determined by Scharber's model, utilizing either the complete dataset or subsets of CEPDB~\cite{lopez2017design, eibeck2021predicting, pyzer2015learning, sun2019use, kong2023prediction}. 
% Various machine learning models, such as K-nearest neighbor (KNN), Random Forest (RF), and Support Vector Regression (SVR), have been applied, using molecular structural descriptions derived from SMILES strings and fingerprint descriptions as inputs. These approaches have effectively captured the relationship between PCE and OSC donor molecules, achieving notable R$^2$ scores, with the highest reaching 0.68~\cite{}. 
Neural network architectures, including Artificial Neural Networks (ANN), Convolutional Neural Networks (CNN), Recurrent Neural Networks (RNN), and Graph Neural Networks (GNN), have demonstrated superior capability in learning from large datasets like CEPDB, yielding an impressive maximum R$^2$ score of 0.996~\cite{eibeck2021predicting}.

\begin{figure*}
    \centering
     \includegraphics[width=0.93\textwidth]{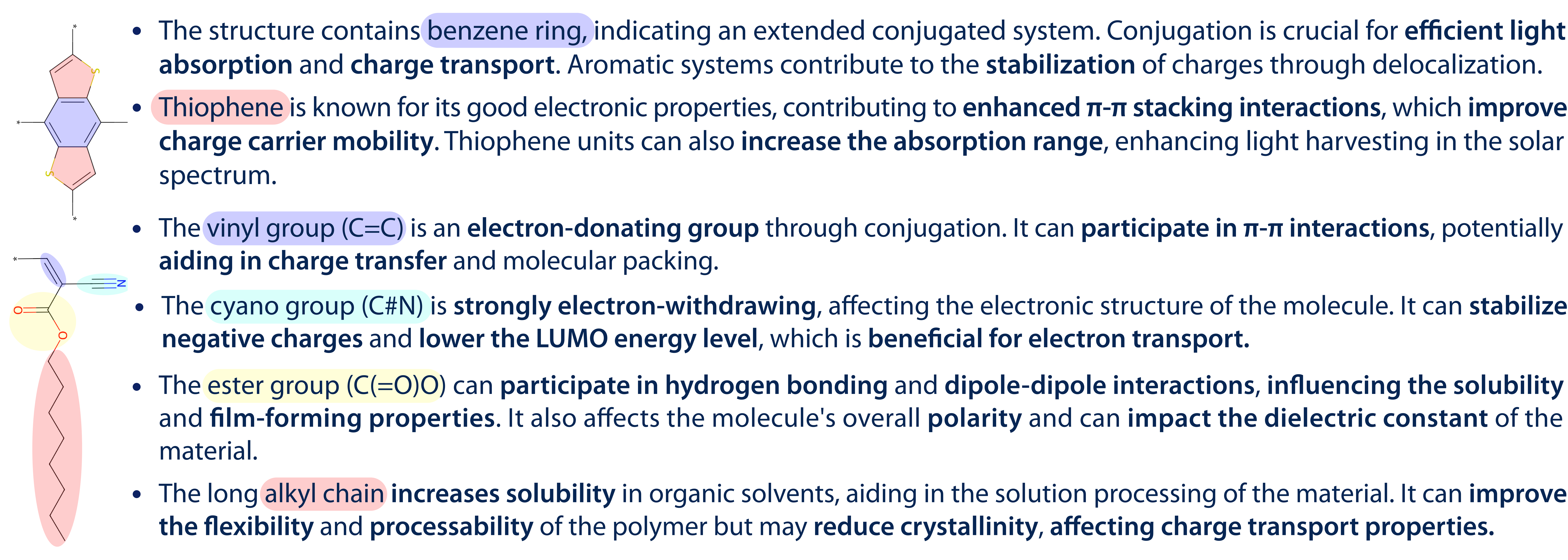}
    \caption{Examples of functional module descriptions  generated by ChatGPT 3.5.}
    \label{fig:text}
\end{figure*}

While some studies achieve high predictive accuracy on computational datasets, less attention is paid to the suitability of these datasets and their agreement with experimental PCE measurements. In a previous work \cite{eibeck2021predicting}, the authors investigate the impact of training data choice and conclude that while current ML models perform well on large computational datasets like CEPDB, fitting on smaller and experimental datasets proves challenging due to numerous degrees of freedom, such as experimental setups and minor device design factors. Moreover, discrepancies between computational PCE based on Scharber’s model and experimental PCE are noted~\cite{greenstein2022computational, hachmann2011harvard}, prompting efforts to collect new OPV datasets~\cite{padula2019combining, wang2023efficient, greenstein2023screening, wu2020machine}. Notably, \textit{Greenstein et al.} constructed a new dataset comprising 1001 unique donor/non-fullerene acceptor pairs, and an ensemble of random forest and neural network models predicting PCE achieves an R$^2$ of 0.4~\cite{greenstein2023screening}.

Our work extends the research on PCE prediction for OPV devices by gathering up-to-date OPV data from the literature and developing a novel predictive model. Unlike previous studies focusing solely on donor molecules~\cite{hachmann2011harvard, lopez2016harvard}, our dataset encompasses a diverse range of donor and acceptor molecules. Each molecule is decomposed into functional modules, and an Attentive FP~\cite{xiong2019pushing} model is employed to extract structural features from each functional module, complemented by textual descriptions generated by an LLM. This approach yields a multimodal dataset providing both structural and property knowledge of molecular functional modules, enabling precise PCE prediction with an R$^2$ of 0.747 ($\pm 0.04$). This method also facilitates modular synthesis of new OPV molecules and sheds light on the relationship between molecular structure and PCE of OPV devices.
% \Tiara{I am not sure we should claim it facilitates modular synthesis. Possibly rephrase: "This method leverages modular synthesis of new OPV molecules..." This would imply we are taking modularity into consideration, but not exactly making this process easier. }
% \thao{Compare to approaches that directly extract features from atoms, our approach provides understandings of the building blocks, thus it can "facilitates modular synthesis"}
\subsection{Multimodal Representation of Molecules: Graph Structure and Textual Descriptions}

% Large language models (LLMs) have demonstrated remarkable capabilities in understanding chemical text and Simplified Molecular Input Line Entry System (SMILES) strings [citations]. These models, trained on vast amounts of SMILES strings and chemical literature, can effectively understand and generate textual descriptions of molecules properties. This proficiency in processing chemical text opens avenues for integrating textual information into molecular representation learning frameworks, complementing graph-based approaches and enriching the understanding of molecular properties and behaviors.

LLMs have emerged as powerful tools for molecular captioning, even from SMILES strings—compact textual representations of molecular structures~\cite{edwards2022translation, liu2023molca, guo2023can}. Models like GPT (Generative Pre-trained Transformer)~\cite{radford2018improving} variants can analyze these strings, generating detailed textual descriptions of molecules. Through fine-tuning on large chemical text datasets, LLMs become proficient at understanding molecular structures encoded in SMILES strings and producing coherent captions~\cite{edwards2022translation}. In this study, we harness the capacity of LLMs to generate structural, physical, chemical, and photovoltaic descriptions of functional modules commonly found in OPV molecules. This allows us to furnish insights into molecular properties that may not be apparent in a molecular graph without background contextual information.  Additionally, this method enhances the factual correctness of the generated text, given the relative ease with which LLMs generate captions for shorter SMILES strings (molecular substructures) compared to longer ones (the entire molecule). We note that functional modules are molecular subgraphs often referred to as fragments in other work \cite{ertl2009estimation}.
% Combining the strengths of graph-based molecular representation and LLMs' proficiency in understanding and generating chemical text, researchers are exploring the potential of multimodal representation learning for molecules. By fusing information from both modalities—molecular graph structure and textual descriptions—into a unified representation, a more comprehensive and holistic understanding of molecules can be achieved. This multimodal approach not only enhances the richness of molecular representations but also facilitates tasks such as molecular similarity comparison, property prediction, and drug discovery. Ultimately, leveraging the synergies between graph-based structures and textual descriptions paves the way for more effective and insightful computational studies in chemistry and chemical informatics.

Several previous studies have focused on incorporating SMILES strings and textual descriptions to enhance molecular understanding tasks. In earlier works \cite{zeng2022deep, liu2023molxpt}, a unified representation of text and SMILES was created by replacing chemical compound names in text with SMILES strings. Other studies \cite{su2022molecular, liu2023multi, cao2023instructmol, liu2023git, zhao2024gimlet} aligned SMILES strings and textual descriptions through contrastive learning or cross-modal projection to ensure that their representations are close in the representation space. Both methods achieved high performance in molecular understanding tasks, with MolXPT~\cite{liu2023molxpt} achieving state-of-the-art results in MoleculeNet tasks~\cite{wu2018moleculenet}.
\begin{figure}[t]
    \centering
    \includegraphics[width=\columnwidth]{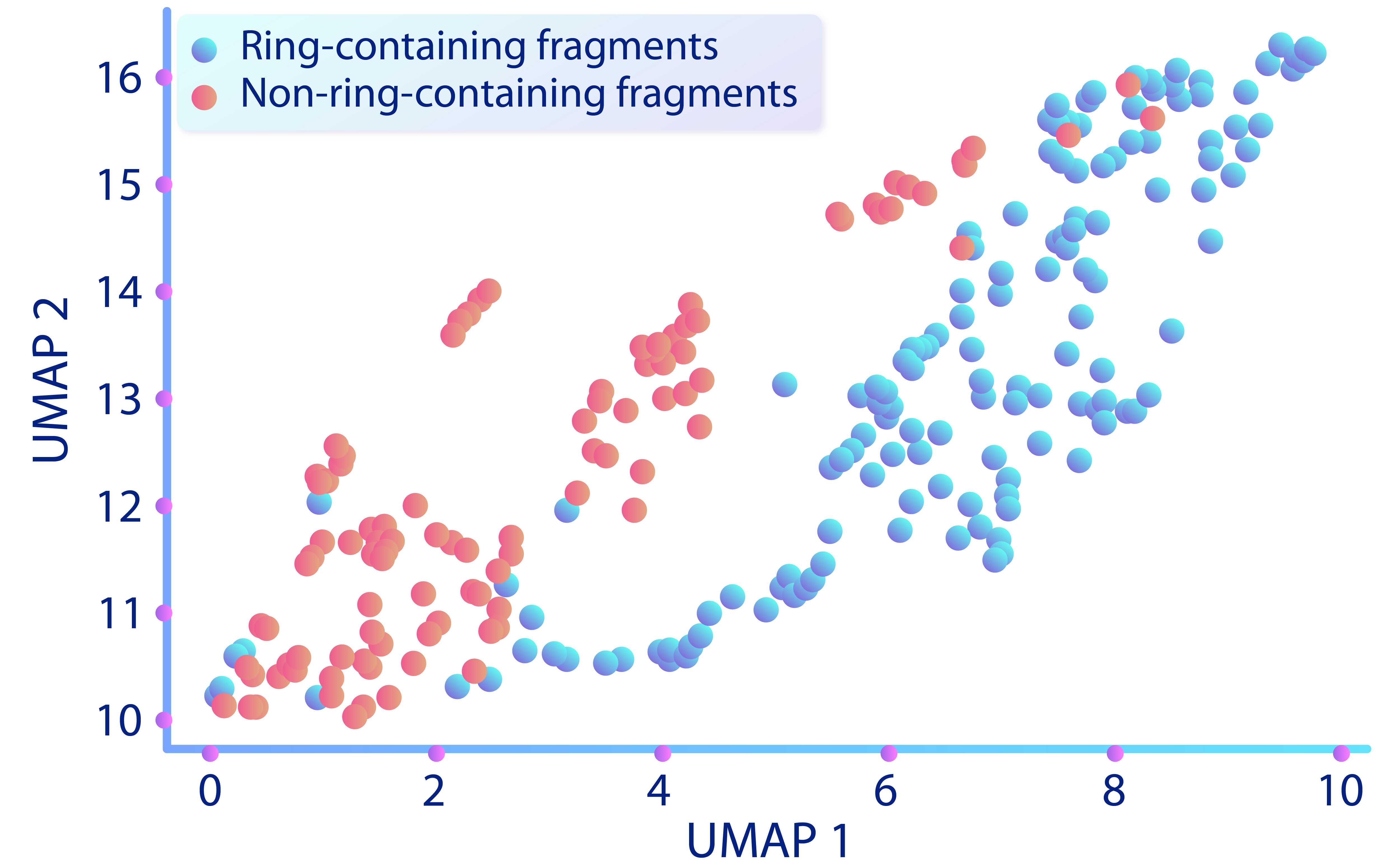}
    \caption{UMAP representation of text descriptions for ring-containing and non-ring-containing fragments in the text embedding space.}
    \label{te_full}
\end{figure}
\begin{figure}[t]
    \centering
    \includegraphics[width=\columnwidth]{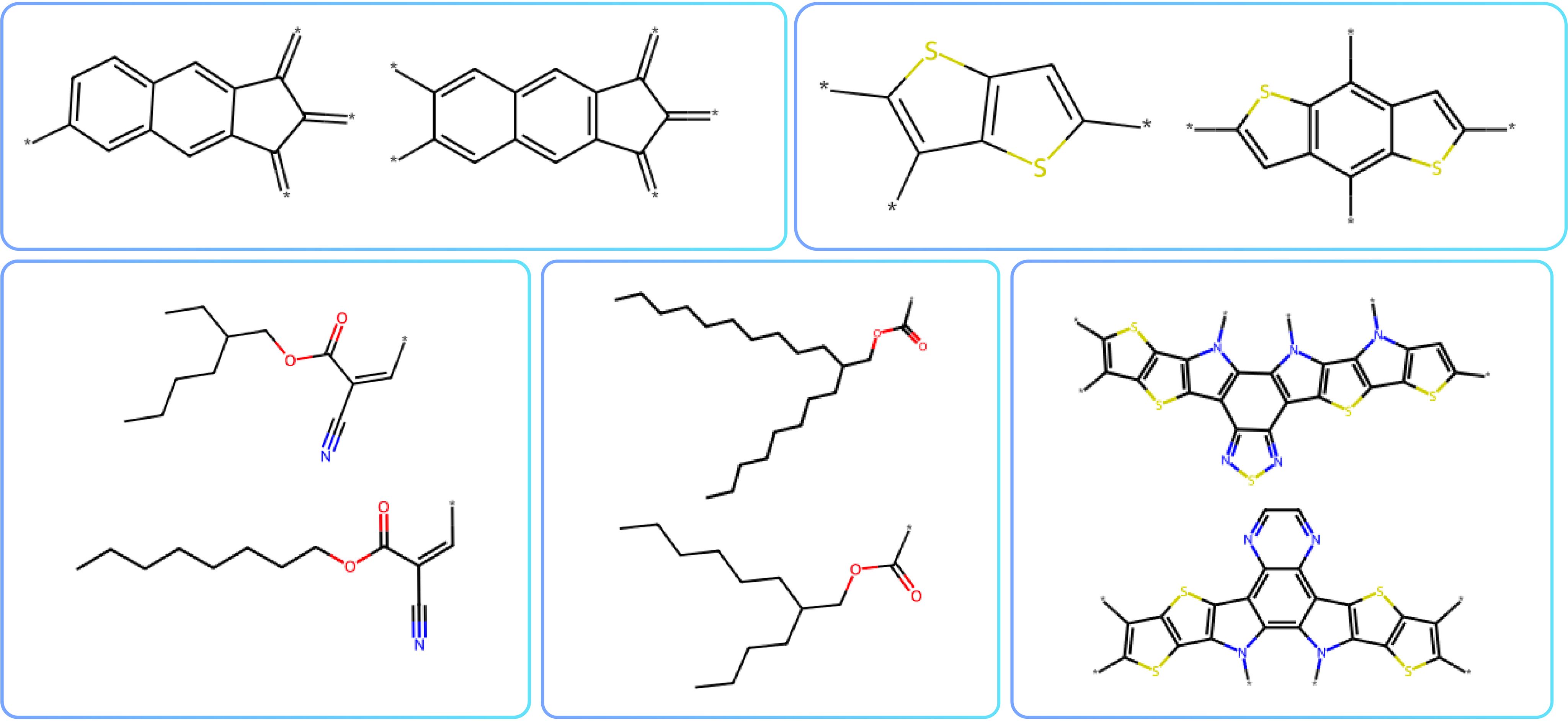}
    \caption{Pairs of fragments that are close in the text embedding space, indicating shared properties or functions. Fragments with similar structures, such as side chains containing ester groups (C(=O)O) and cyano groups (C\#N), are clustered together in the embedding space, reflecting their structural similarities.}
    \label{blocks}
\end{figure}

While several studies have combined knowledge graphs and text descriptions to enhance the representations of either modality or both~\cite{jin2023large}, no prior research has integrated textual data into graphs of molecular fragments (functional modules). In this study, for the first time, we integrate structural embeddings obtained from a GNN model and text embeddings obtained from LLMs to form a multimodal representation of such functional modules. By doing so, our model can make predictions based on information from both modalities, ultimately enhancing its performance on a wide variety of prediction tasks.

\section{OPV Dataset Collection}
% We collected an OPV dataset comprising donor-acceptor pairs used in OPV devices, along with their associated properties. In order to complement the molecular structural descriptors, we generated textual descriptions for the fragments that constitute the molecules in our dataset. This section provides details of the collected dataset, in terms of molecular data and text data.

% \subsection{Collect OPV Data from Literature}
Due to the lack of curated high-quality experimental data, we curated an OPV dataset to train our PCE prediction model. The dataset consists of 500 pairs of donor and acceptor molecules employed in bulk heterojunction (BHJ) and bilayer OPV devices collected from literature from 2012 to 2023. 

In this dataset, there are a total of 403 molecular entities (comprised by 10 atoms: C, H, O, N, S, Si, Se, Cl, Br, F), including 203 donor molecules and 252 acceptor molecules (with 52 molecules that can be either donor or acceptor in a device). It includes properties of OPV devices such as PCE, open circuit potential (V$_{oc}$), short circuit current density (J$_{sc}$), and fill factor (FF) for each donor-acceptor pair.
Table~\ref{table:data_stats} provides the statistics of the collected dataset.
Compared to the HOPV dataset~\cite{lopez2016harvard}, our dataset demonstrates superior diversity, encompassing a significantly larger portion of the chemical space. We attribute this to five key differences: 

\begin{enumerate}
    \item It contains pairs of donor-acceptor molecules, instead of solely donor molecules as in HOPV;
    \item It includes up-to-date data of OPV devices with a higher PCE range, from 2.5\% to 19.6\%, compared to 0.0005\% to 10.2\% in HOPV;
    \item It comprises molecules of greater diversity, reflected in a lower average Tanimoto distance~\cite{bajusz2015tanimoto} of 0.67, compared to 0.8 in HOPV (refer to Figure~\ref{fig:tanimoto} and Figure~\ref{fig:hopv_tanimoto} to observe the difference in Tanimoto distance between the collected dataset and HOPV);
    \item It contains a more diverse range of atom types (10 atoms) compared to the 8 atom types present in HOPV (C, H, O, N, F, S, Si, and Se);
    \item It contains a larger number of samples (500 samples compared to 350 in HOPV).
\end{enumerate}

Each molecule in the dataset is further decomposed into functional modules, also referred to as fragments, for additional processing. A total of 250 different functional modules result from the decomposition of the 403 molecules in the dataset.
% Table~\ref{table:data_stats} contains detailed statistical information about the collected dataset.

With this dataset, our objective is to construct machine learning models capable of accurately predicting the PCE score based on pairs of donor and acceptor molecules. A robust PCE prediction model is characterized by a high coefficient of determination (R$^2$), low Mean Square Error (MSE), and low Mean Absolute Error (MAE).
\begin{figure}[t]
    \centering
    \includegraphics[width=0.35\textwidth]{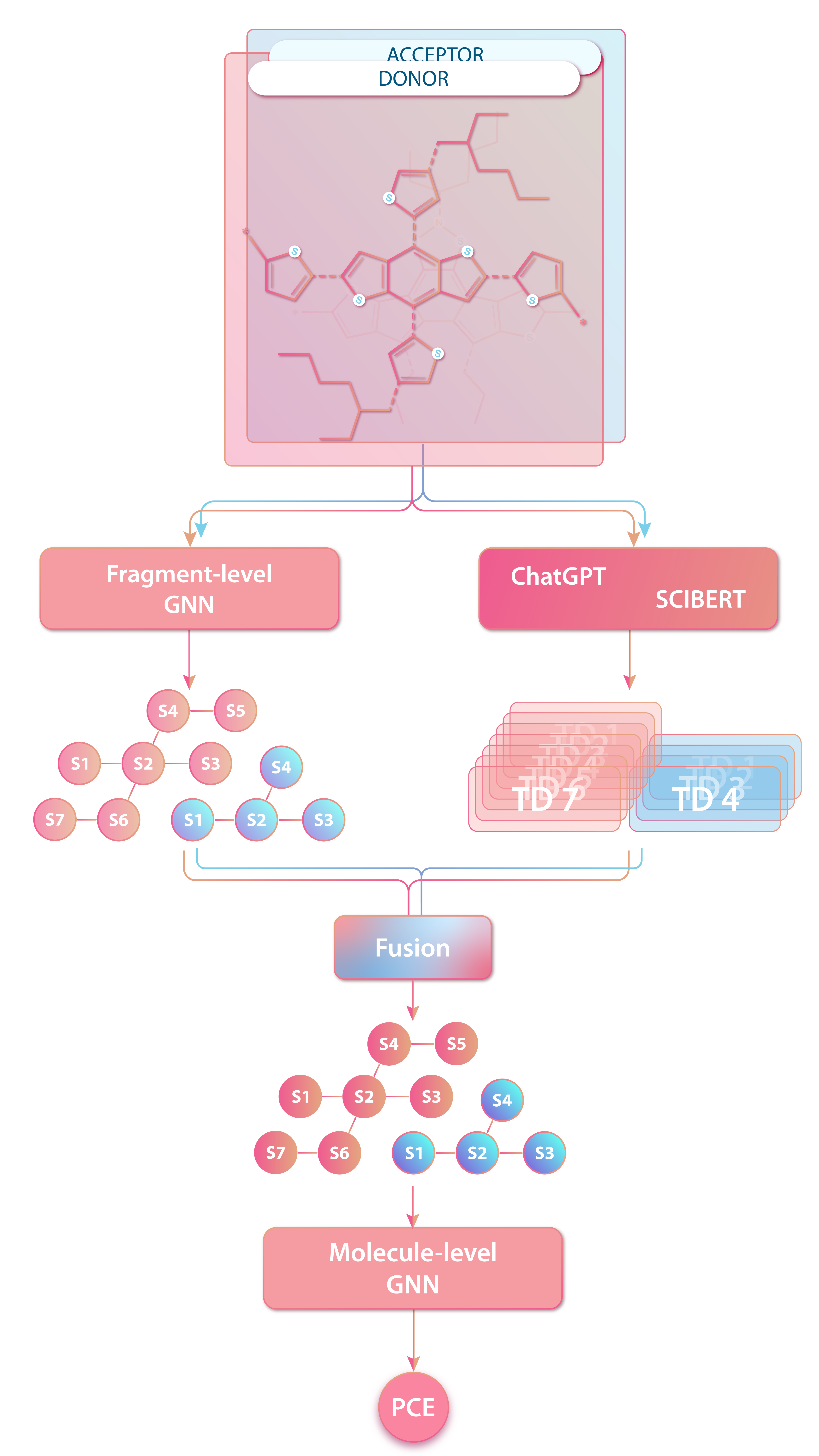}
    \caption{Proposed model architecture (TD: Textual descriptors of a functional module).}
    \label{main_diagram}
\end{figure}

\begin{table}[t]
    \centering
    \small
    \setlength{\tabcolsep}{3pt}
    \resizebox{0.98\columnwidth}{!}{
    \begin{tabular}{ccccc}
        \toprule
        & \textbf{\makecell{Hierarchical \\GNN}} & \textbf{\makecell{Molecule- \\ Level \\ GNN}} & \textbf{\makecell{Donor-only \\GNN}} & \textbf{\makecell{Acceptor-only \\GNN}} \\
        \midrule
%        \multirow{2}{*}{\textbf{MSE}} & \textbf{Avg} & \textbf{3.583} & 4.612 & 10.178 & 6.648 \\
        \textbf{MSE} & \textbf{3.58} & 4.61 & 10.18 & 6.65 \\
%        & \makecell{95\%\\ CI} & \makecell{[2.81, \\4.412]} & \makecell{[3.818, \\ 5.314]} & \makecell{[8.689, \\ 11.792]} & \makecell{[5.669, \\ 7.412]} \\
        \makecell{95\% CI} & \makecell{[2.81, 4.41]} & \makecell{[3.82, 5.31]} & \makecell{[8.69, 11.79]} & \makecell{[5.67, 7.41]} \\
        \midrule%\cmidrule{2-6}
        \textbf{MAE} & \textbf{1.461} & 1.792 & 2.726 & 2.162 \\
%        & \makecell{95\%\\ CI} & \makecell{[1.282, \\ 1.664]} & \makecell{[1.621, \\ 1.947]} & \makecell{[2.453, \\ 2.952]} & \makecell{[1.988, \\ 2.326]} \\
         \makecell{95\% CI} & \makecell{[1.28, 1.66]} & \makecell{[1.62, 1.95]} & \makecell{[2.45, 2.95]} & \makecell{[1.99, 2.33]} \\
        \midrule%\cmidrule{2-6}
%        \multirow{2}{*}{\textbf{R$^2$}} & \textbf{Avg} & \textbf{0.644} & 0.534 & 0.398 & 0.428 \\
        \textbf{R$^2$} & \textbf{0.644} & 0.534 & 0.398 & 0.428 \\
%        & \makecell{95\%\\ CI} & \makecell{[0.598, \\ 0.687]} & \makecell{[0.48, \\ 0.591]} & \makecell{[0.342, \\ 0.451]} & \makecell{[0.381, \\ 0.480]} \\
        \makecell{95\% CI} & \makecell{[0.59, 0.68]} & \makecell{[0.48, 0.59]} & \makecell{[0.34, 0.45]} & \makecell{[0.38, 0.48]} \\
        \bottomrule
        
    \end{tabular}}
    \caption{PCE prediction results on the collected dataset with different GNN architectures (average of 30 runs).}
    \label{table:hierarchicalGNN}
\end{table}

\begin{table*}[t]
    % \caption{PCE prediction results on the collected dataset of the baseline model (hierarchical GNN without incorporating textual descriptors) and models incorporated with different kinds of textual descriptors.}
    % \begin{tabular}{p{0.04\textwidth} p{0.07\textwidth} p{0.1\textwidth} p{0.1\textwidth} p{0.1\textwidth} p{0.1\textwidth} p{0.1\textwidth} p{0.1\textwidth} p{0.1\textwidth}}
    \centering
    \small
    \renewcommand{\arraystretch}{1.12}
    \setlength{\tabcolsep}{3pt}
    \resizebox{0.99\textwidth}{!}{
    \begin{tabular}{lllllllll}
        \toprule
        & & \textbf{Baseline} & \makecell{\textbf{Full} \\ \textbf{description}} & \textbf{Structural} & \textbf{Physical} & \textbf{Chemical} & \textbf{Photovoltaic} & \makecell[c]{\textbf{Physical+} \\ \textbf{Chemical}} \\
        \midrule
        \multirow{2}{*}{\textbf{MSE}} & \textbf{Avg} & 3.583 & 2.878 & 2.66 & 2.561 & 2.308 & 3.317 & 2.327 \\
        & 95\% CI & [2.81, 4.412] & [1.282, 3.51] & [1.445, 4.283] & [1.883, 3.322] & [1.851, 2.858] & [2.579, 4.176] & [1.584, 3.186] \\
        \hline
        \multirow{2}{*}{\textbf{MAE}} & \textbf{Avg} & 1.461 & 1.32 & 1.289 & 1.231 & 1.218 & 1.367 & 1.194 \\
        & 95\% CI & [1.282, 1.664] & [1.174, 1.482] & [1.053, 1.563] & [1.059, 1.42] & [1.08, 1.366] & [1.171, 1.56] & [0.985, 1.479] \\
        \hline
        \multirow{3}{*}{\textbf{R$^2$}} & \textbf{Avg}  & 0.644 & 0.703 & 0.725 & 0.732 & 0.735 & 0.659 & \textbf{0.747} \\
        & ($\uparrow$) & - & $\uparrow$ 0.059 & $\uparrow$ 0.081 & $\uparrow$ 0.088 & $\uparrow$ 0.091 & $\uparrow$ 0.015 & $\uparrow$ \textbf{0.103} \\
        & 95\% CI & [0.658, 0.757] & [0.688, 0.759] & [0.694, 0.774] & [0.688, 0.779] & [0.611, 0.703] & [0.703, 0.794] & [0.698, 0.787] \\
        \bottomrule
        
    \end{tabular}}
    \caption{PCE prediction results on the collected dataset of the baseline model (hierarchical GNN without incorporating textual descriptors) and models incorporated with different kinds of textual descriptors.}
    \label{table:collected_data_results}
\end{table*}

\section{Fusing Text with Molecular Structure}%Incorporating Text and Structural Descriptors}

In this section, we detail our approach, \model{}, for extracting structural and textual descriptors for each functional module, and then fusing them to form the multimodal representation of those modules. Figure~\ref{main_diagram} illustrates the architecture of our proposed model.

\begin{table}[t]
    \centering
    \footnotesize
    \begin{tabular}{>{\centering}p{0.04\textwidth} >{\centering}p{0.04\textwidth} >{\centering}p{0.07\textwidth} >{\centering}p{0.07\textwidth} p{0.07\textwidth}}
    % \setlength{\tabcolsep}{3pt}
    % \resizebox{0.29\textwidth}{!}{
    % \begin{tabular}{ccccc}
        \toprule
        & & \textbf{w/o text} & \textbf{w/ text} & \textbf{SVR}\\
        \midrule
        \multirow{2}{*}{\textbf{MSE}} & \textbf{Avg} & 2.598 & \textbf{2.321} & 2.687\\
        & $\pm$ & 0.524 & 0.487 & 0.487\\
        \multirow{2}{*}{\textbf{MAE}} & \textbf{Avg} & 1.233 & \textbf{1.034} & 1.132\\
        & $\pm$ & 0.146 & 0.136 & 0.095\\
        \multirow{2}{*}{\textbf{R$^2$}} & \textbf{Avg} & 0.492 & \textbf{0.588} & 0.453\\
        & $\pm$ & 0.109 & 0.115 & 0.109\\
        \bottomrule
        
    \end{tabular}
    \caption{PCE prediction results on the HOPV dataset of the proposed method and SVR model \cite{eibeck2021predicting}}
    \label{table:hopv}
\end{table}

\subsection{Modeling Molecular Structure}

    After collecting the SMILES strings of OPV molecules, we construct molecular graphs and employ a molecular decomposition algorithm to decompose them into constituent functional modules. This algorithm breaks down molecules at C-C single bonds between conjugated backbone rings and their corresponding side chains. This approach harnesses modular synthesis, wherein complex molecules are iteratively assembled from smaller constituent functional modules~\cite{li2015synthesis, gillis2009iterative, blair2022synthesis}.

    Fragment-level graphs representing functional modules will undergo processing by a GNN model to produce structural descriptors. Various GNN architectures, including Graph Convolutional Networks (GCN)~\cite{kipf2016semi}, Graph Attention Networks (GAT)~\cite{velivckovic2017graph}, and Attentive FP~\cite{xiong2019pushing}, are employed to extract structural descriptors from molecular graphs. Subsequently, these structural descriptors of each functional module will be fused with textual descriptors to create a multimodal representation of each functional module.
    
\subsection{Generating Textual Descriptions for Functional Modules}
For each functional module, ChatGPT-3.5~\citep{gpt3.5} is utilized to generate descriptions including their structural, physical, chemical, and photovoltaic properties. A total of 250 descriptions are produced. These descriptions then undergo manual evaluation to ensure the factual accuracy of the generated text. A subset of 60  functional modules and their descriptions is manually evaluated, revealing that 88\% (53 out of 60) are correct. Figure~\ref{fig:text} exemplifies a result generated by GPT-3.5.

% To generate text descriptions for functional modules, we use their SMILES string to query ChatGPT-3.5~\cite{gpt3.5} with this prompt:
% \textit{Generate descriptions of  this molecular fragment:} ${\mathsf[SMILES]}$ \textit{focusing on its structural, physical, chemical, and photovoltaic properties. Descriptions should be specific and tailored for organic photovoltaic (OPV) material research. Avoid neutral information.}

%\subsubsection{Modeling Textual Descriptions}

\subsection{Modeling Textual Descriptions}

Textual descriptions of functional modules are fed into a frozen Scibert~\cite{beltagy2019scibert} model to extract text embeddings. We assessed the efficacy of combining descriptions for each property with structural descriptors to identify those yielding improvements, which will be retained for the generation of textual descriptors.% for fragments.

In order to evaluate the quality of textual descriptors, Figure~\ref{te_full} illustrates the UMAP~\cite{mcinnes1802umap} representation of descriptors for fragments containing rings and those without rings. The two groups exhibit distinguishable patterns, indicating that text generated by ChatGPT 3.5 can differentiate between side chain blocks and ring blocks. Furthermore, we randomly selected data points that are close to each other in the text embedding space. Outcomes (depicted in Figure~\ref{blocks}) show that functional modules with similar structures tend to cluster together in the text embedding space, suggesting that textual descriptors effectively capture information regarding the similarity of molecular fragments. 
    % \begin{itemize}
    %     \item SMILES of fragments $\rightarrow$ GPT 3.5 $\rightarrow$ textual description $\rightarrow$ Scibert $\rightarrow$ textual descriptors
    %     \item Ablation study: different kinds of description: Structure, Physical, Chemical, Photovoltaic
    % \end{itemize}

\begin{table*}[t]
    \centering
    \footnotesize
    \renewcommand{\arraystretch}{1.2}
        % \begin{tabular}{p{0.03\textwidth} p{0.03\textwidth} p{0.08\textwidth} p{0.08\textwidth} p{0.08\textwidth} p{0.08\textwidth} p{0.08\textwidth} p{0.08\textwidth} p{0.08\textwidth} p{0.08\textwidth}}
        \setlength{\tabcolsep}{3pt}
        \resizebox{0.7\textwidth}{!}{
        \begin{tabular}{llcccccccc}
        \toprule
        & & \multicolumn{2}{c}{\textbf{B3LYP}} & \multicolumn{2}{c}{\textbf{BP86}} & \multicolumn{2}{c}{\textbf{M06-2X}} & \multicolumn{2}{c}{\textbf{PBE0}} \\
        & & \textbf{w/o text} & \textbf{w/ text} & \textbf{w/o text} & \textbf{w/ text} & \textbf{w/o text} & \textbf{w/ text} & \textbf{w/o text} & \textbf{w/ text} \\
        \toprule
        \multirow{2}{*}{\textbf{MSE}} & \textbf{Avg} & 0.064 & 0.035 & 3.487 & 0.188 & 2e-4 & 1e-4 & 0.036 & 0.003 \\
        & $\pm$ & 0.022 & 0.01 & 1.185 & 0.064 & 8e-5 & 2e-5 & 0.011 & 6e-4 \\
        \cline{3-10}
        \multirow{2}{*}{\textbf{MAE}} & \textbf{Avg} & 0.182 & 0.136 & 1.354 & 0.273 & 0.031 & 0.005 & 0.133 & 0.038 \\
        & $\pm$ & 0.025 & 0.014 & 0.191 & 0.035 & 0.002 & 0.001 & 0.019 & 0.004 \\
        % \hline
        \cline{3-10}
        \multirow{2}{*}{\textbf{R$^2$}} & \textbf{Avg} & 0.943 & \textbf{0.968} & 0.935 & \textbf{0.964} & 0.966 & \textbf{0.974} & 0.951 & \textbf{0.996} \\
        & $\pm$ & 0.02 & 0.019 & 0.018 & 0.02 & 0.019 & 0.021 & 0.017 & 8e-4 \\
        \toprule
        
    \end{tabular}}
    \caption{Results of predicting computational PCE on the HOPV dataset, using computational PCE obtained from Scharber's model with a selection of four functionals (B3LYP, BP86, M06-2X, and PBE0).}
    \label{table:hopv_dft}
\end{table*}

\begin{table*}[t]
    \centering
    \footnotesize
    \setlength{\tabcolsep}{3pt}
    \resizebox{0.7\textwidth}{!}{
    \begin{tabular}{lccccc}
        % \begin{tabular}{p{0.2\textwidth} p{0.1\textwidth} p{0.1\textwidth} p{0.1\textwidth} p{0.1\textwidth}}
        \toprule
         \textbf{Dataset}& \textbf{BBBP} & \textbf{BACE} & \textbf{ClinTox} & \textbf{SIDER} \\
         \textbf{\#molecules} & 2039 & 1513 & 1478 & 1427 \\
         \textbf{\#tasks} & 1 & 1 & 2 & 27 \\
         \midrule
        % \textbf{G-Motif}~\cite{} & 66.4 (1.6) & 73.4 (4.0) & 77.8 (2.0) & 58.4 (0.6) \\ 
        \textbf{D-MPNN}~\cite{yang2019analyzing} & 71.0 (0.3) & 80.9 (0.6) & 90.6 (0.6) & 57.0 (0.7) \\
        \textbf{AttentiveFP}~\cite{xiong2019pushing} & 64.3 (1.8) & 78.4 (0.02) & 84.7 (0.3) & 60.6 (3.2) \\
        \textbf{GROVER}~\cite{rong2020self} & 69.5 (0.1) & 81.0 (1.4) & 76.2 (3.7) & 65.4 (0.1)  \\
        \textbf{MolCLR}~\cite{wang2022molecular} & 72.2 (2.1) & 82.4 (0.9) & \textbf{91.2} (3.5) & 58.9 (1.4) \\
        \textbf{GraphMVP}~\cite{liu2021pre} & 72.4 (2.1) & 81.2 (0.9) & 79.1 (2.8) & 63.9 (1.2) \\
        \textbf{GEM}~\cite{fang2022geometry} & 72.4 (0.4) & 85.6 (1.1) & 90.1 (1.3) &  67.2 (0.4)\\
        \textbf{HiMo$_{small}$}~\cite{zang2023hierarchical} & 71.3 (0.6) & 84.6 (0.2) & 70.6 (2.1) & 62.5 (0.3) \\
        \textbf{HiMo$_{large}$}~\cite{zang2023hierarchical} & 73.2 (0.8) & 84.3 (0.3) & 80.8 (1.4) & 61.3 (0.5) \\
        % \textbf{UniMol} & 72.9 (0.6) & 85.7 (1.2) & 91.9 (1.8) & 80.8 (0.3) \\
        % \hline
        % \textbf{KV-PLM}~\cite{zeng2022deep} & 74.6 (0.9) & - & - & 61.5 (1.5) \\
        % % \textbf{Galactica}~\cite{taylor2022galactica} & 66.1 & 61.7 & 82.6 & 63.2 \\
        % \textbf{MoMu}~\cite{su2022molecular} & 70.5 (2.0) & 77.1 (1.4) & 79.9 (4.1) & 60.5 (0.9) \\
        % \textbf{MolXPT} & 80.0 (0.5) & \textbf{88.4} (1.0) & \textbf{95.3} (0.2) & 71.7 (0.2) \\
        
        \hline
        \model{} \textbf{(w/o text)} & 82.8 (1.2) & 82.1 (0.8) & 85.6 (1.7) & 64.3 (0.9) \\
        \model{} \textbf{(w/ text)} & \textbf{86.4} (1.5) & \textbf{85.7} (0.9) & 87.3 (1.2) & \textbf{68.1} (1.3) \\
        \bottomrule
        
    \end{tabular}}
    \caption{Results of \model{} on four common MoleculeNet tasks compared to other GNN-based models. Evaluation metric: ROC-AUC(\%)}
    \label{table:molnet}
\end{table*}

\subsection{Fusion Approaches}
    
    After generating both structural and textual descriptors for functional modules, we combine them using fusion operators. We evaluate two fusion operators: \textit{average + concat} and \textit{attention + concat}.

    The first approach computes an embedding for the entire text description by averaging all the word embeddings, and then concatenates this with the structural embedding to form a multimodal representation of the functional module, denoted as~$\mathbf{v}$.
    
    The attention-based module comprises learnable query ($\mathbf{W}_Q$), key ($\mathbf{W}_K$), and value ($\mathbf{W}_V$) matrices to learn the cross-attention score between the structural embedding vector $\mathbf{s}$ of a functional module and the word embedding vectors $\mathbf{t}$ of its description. The attention weight is calculated by Equation~\ref{eq: attention}.
    \begin{equation}
    \alpha = \text{softmax}\left(\frac{\mathbf{Q}\mathbf{K}^T}{\sqrt{d_k}}\right)\mathbf{V}
    \label{eq: attention}
    \end{equation}
    Here, $\alpha$ represents the cross-attention score between $\mathbf{s}$ and $\mathbf{t}$, where $\mathbf{Q} = \mathbf{W}_Q \cdot \mathbf{t}$, $\mathbf{K} = \mathbf{W}_K \cdot \mathbf{s}$, and $\mathbf{V} = \mathbf{W}_V \cdot \mathbf{s}$. The term $d_k$ denotes the dimensionality of the key $\mathbf{s}$.

    The embedding of the entire text description is computed as the weighted average of word embeddings, with the attention scores from the structural embedding serving as the weights, as shown in Equation~\ref{eq: attention1}.
    \begin{equation}
        \mathbf{t'} = \sum_{i=0}^{N}\alpha_i \cdot \mathbf{t}_i
        \label{eq: attention1}
    \end{equation}
        Here $N$ represents the length of the text description.

    Finally, the structural and textual embeddings are concatenated to create a multimodal representation of each functional module $\mathbf{v}$, expressed as $\mathbf{v} = \text{concat}(\mathbf{s}, \mathbf{t'})$.
    
    After fusion, each functional module is represented by a vector $\mathbf{v}$, representing a node in the molecule-level graph. The edges of this graph are defined by the bonds connecting the functional modules. This graph is input to the molecule-level GNN model, which outputs a predicted PCE score for the input donor-acceptor pair.
    %Figure~\ref{main_diagram} illustrates the architecture of our proposed model \model{}.
    
\section{Experiments}
\subsection{Experimental Settings}
    \textbf{Datasets and evaluation metrics.} We conduct an evaluation of \model{} across multiple datasets, including our collected OPV dataset, the HOPV dataset~\cite{lopez2016harvard}, and several tasks from the MoleculeNet benchmark dataset (BBBP, BACE, ClinTox, and SIDER)~\cite{wu2018moleculenet}.
    To evaluate the efficacy of our proposed method across both computational and experimental data, we assess its performance on the HOPV dataset for experimental PCE and PCE computed using Scharber's model~\cite{scharber2006design}. We utilize three commonly used metrics: R$^2$, MSE, and MAE for PCE prediction task. Meanwhile, for the MoleculeNet tasks, we employ the AUC-ROC metric to evaluate its performance.
    
    \textbf{Data split.} 
    In accordance with the experimental setups in previous work~\cite{eibeck2021predicting}, we split the two OPV datasets into training, validation, and test sets with a ratio of 80:10:10. Similarly, for the MoleculeNet tasks, we split the dataset into training, validation, and test sets with ratio of 80:10:10, respectively, following previous research~\cite{zhou2023uni}.
    
    \model{}\textbf{'s model architecture.}
    We evaluate various model architectures by experimenting with the following setups, testing their performance on the collected dataset:
    \begin{itemize}
        \item Different GNNs (including a GNN that takes a molecular graph as input and directly outputs predictions of PCE, versus a hierarchical GNN including a fragment-level GNN that extracts structural descriptors of functional modules followed by a molecule-level GNN that takes multimodal representations of fragments as input);
        \item Different kinds of textual descriptions (structural, physical, chemical, photovoltaic property descriptions, and descriptions of all properties).
    \end{itemize}
% \begin{itemize}
%     \item Experiments
%     \item Experimental setups
%     \item Experimental results
%         \begin{itemize}
%         \item Main results: collected data, baseline + different kinds of description
%         \item Result on HOPV: results on experimental data, results on computational data
%         \item Result on MoleculeNet: HIV, Tox21, bbbp, SIDER
%         \end{itemize}
% \end{itemize}

\subsection{Main Results}
\textbf{Results on the collected dataset:} The results of using a molecule-level GNN and a hierarchical GNN (without text) are described in Table~\ref{table:hierarchicalGNN}. According to our findings, using a hierarchical GNN architecture that combines fragment-level and molecule-level GNNs results in a significant improvement in the R$^2$ score of 0.11 ($\pm$ 0.04) when compared to using only molecule-level GNN. We also examine the effectiveness of using only donor or acceptor molecules as input for the hierarchical GNN model. We find that R$^2$ score can be greatly increased by using pairs of donor and acceptor molecules as input. This leads to an R$^2$ score of 0.644 ($\pm$ 0.05), whereas models that use either only donor or only acceptor molecules have R$^2$ of 0.398 and 0.428, respectively.\\ 
Table~\ref{table:collected_data_results} shows experimental results of incorporating various kinds of textual descriptors with structural descriptors obtain from the fragment-level GNN. We observe that using textual description of all properties improve predictive performance from 0.015 to 0.103 in R$^2$ score, with the highest improvement from physical and chemical descriptions, and the combination of both.

\textbf{Results on the HOPV dataset.} The results from experiments employing the proposed \model{} model on the HOPV dataset are presented in Table~\ref{table:hopv} and Table~\ref{table:hopv_dft}. These results demonstrate that \model{} outperforms another method using the SVR model by 0.135 in R$^2$ score in the task of predicting experimental PCE. Table~\ref{table:hopv_dft} also demonstrates \model{}'s ability to accurately predict computational PCE, achieving the highest R$^2$ score of 0.996.
It is worth noting that complementing structural descriptors with textual descriptors consistently improves the predictive performance of the model, both in the collected dataset and HOPV dataset.

\textbf{Results on the MoleculeNet datasets.} Table~\ref{table:molnet} shows the results of using \model{} to solve four classification tasks with small datasets in MoleculeNet~\cite{wu2018moleculenet}. \model{} outperforms other GNN-based models on 3 out of 4 tasks, achieving a significant margin in the BBBP task with a gap of 13.2 ($\pm$ 1.5)\% compared to the second-best method (HiMo~\cite{zang2023hierarchical}). These results demonstrate \model{}'s ability to excel in tasks beyond PCE prediction, proving particularly valuable for low-resource tasks where data collection is challenging. This demonstrates incorporating fragment-level text descriptions can significantly enrich molecule representation.

\section{Conclusion \& Future Work}
In this study, we introduce a new dataset and present a novel approach for predicting PCE in OPV devices. Our approach leverages the learned properties of LLMs to enrich molecular representations at the level of functional modules (molecular fragments). This representation enables accurate prediction of the PCE of OPV devices, as well as other property prediction tasks. However, to apply PCE prediction to high-throughput screening of OPV materials, enhancing prediction reliability is crucial. To achieve this, we plan to incorporate uncertainty quantification methods at both the molecular and functional module levels into our future work. By doing so, we aim to further strengthen our predictions and advance the field of OPV material screening.

\section*{Acknowledgments}
We would like to thank Martin Burke for his helpful discussion. This research is based upon work supported by the Molecule Maker Lab Institute: an AI research institute program supported by NSF under award No. 2019897. The views and conclusions contained herein are those of the authors and should not be interpreted as necessarily representing the official policies, either expressed or implied, of the U.S. Government. The U.S. Government is authorized to reproduce and distribute reprints for governmental purposes notwithstanding any copyright annotation therein.

\section*{Ethics Considerations}
In conducting this study, we ensured ethical compliance by sourcing data from publicly available scientific literature and databases, adhering to standards of data integrity and transparency. Our model, \model{}, is intended to complement human expertise and support decision-making in scientific research, particularly in low-data regimes. We advocate for responsible use of \model{} by validating its predictions with domain experts to ensure safety and reliability in applications such as drug and material discovery. Additionally, we are committed to transparency and reproducibility by sharing our methodologies and findings with the broader scientific community.

\clearpage

% Bibliography entries for the entire Anthology, followed by custom entries
%\bibliography{anthology,custom}
% Custom bibliography entries only
\bibliography{acl_latex}

\begin{thebibliography}{53}
\providecommand{\natexlab}[1]{#1}

\bibitem[{Abdulrazzaq et~al.(2013)Abdulrazzaq, Saini, Bourdo, Dervishi, and Biris}]{abdulrazzaq2013organic}
Omar~A Abdulrazzaq, Viney Saini, Shawn Bourdo, Enkeleda Dervishi, and Alexandru~S Biris. 2013.
\newblock Organic solar cells: a review of materials, limitations, and possibilities for improvement.
\newblock \emph{Particulate science and technology}, 31(5):427--442.

\bibitem[{Adamo and Jacquemin(2013)}]{adamo2013calculations}
Carlo Adamo and Denis Jacquemin. 2013.
\newblock The calculations of excited-state properties with time-dependent density functional theory.
\newblock \emph{Chemical Society Reviews}, 42(3):845--856.

\bibitem[{Bajusz et~al.(2015)Bajusz, R{\'a}cz, and H{\'e}berger}]{bajusz2015tanimoto}
D{\'a}vid Bajusz, Anita R{\'a}cz, and K{\'a}roly H{\'e}berger. 2015.
\newblock Why is tanimoto index an appropriate choice for fingerprint-based similarity calculations?
\newblock \emph{Journal of cheminformatics}, 7:1--13.

\bibitem[{Beltagy et~al.(2019)Beltagy, Lo, and Cohan}]{beltagy2019scibert}
Iz~Beltagy, Kyle Lo, and Arman Cohan. 2019.
\newblock Scibert: A pretrained language model for scientific text.
\newblock \emph{arXiv preprint arXiv:1903.10676}.

\bibitem[{Blair et~al.(2022)Blair, Chitti, Trobe, Kostyra, Haley, Hansen, Ballmer, Woods, Wang, Mubayi, Schmidt, Pipal, Morehouse, Palazzolo~Ray, Gray, Gill, and Burke}]{blair2022synthesis}
Daniel~J. Blair, Chitti Chitti, Sriyankari, Melanie Trobe, David~M Kostyra, Hannah M.~S. Haley, Richard~L. Hansen, Steve~G. Ballmer, Toby~J. Woods, Wesley Wang, Vikram Mubayi, Michael~J. Schmidt, Robert~W. Pipal, Greg~F. Morehouse, Andrea~M. Palazzolo~Ray, Danielle~L. Gray, Adrian~L. Gill, and Martin~D. Burke. 2022.
\newblock Automated iterative csp3–c bond formation.
\newblock \emph{Nature}, 604(7904):92--97.

\bibitem[{Burlingame et~al.(2020)Burlingame, Ball, and Loo}]{burlingame2020stability}
Quinn Burlingame, Melissa Ball, and Yueh-Lin Loo. 2020.
\newblock It's time to focus on organic solar cell stability.
\newblock \emph{Nature Energy}, 5(12):947--949.

\bibitem[{Cao et~al.(2023)Cao, Liu, Lu, Yao, and Li}]{cao2023instructmol}
He~Cao, Zijing Liu, Xingyu Lu, Yuan Yao, and Yu~Li. 2023.
\newblock Instructmol: Multi-modal integration for building a versatile and reliable molecular assistant in drug discovery.
\newblock \emph{arXiv preprint arXiv:2311.16208}.

\bibitem[{Cui et~al.(2021)Cui, Xu, Yao, Bi, Ling, Zhang, Zu, Zhang, Qin, Ren, Chen, He, Hao, Wei, and Hou}]{cui2021opv19}
Yong Cui, Ye~Xu, Huifeng Yao, Pengqing Bi, Hong Ling, Jianqi Zhang, Yunfei Zu, Tao Zhang, Jinzhao Qin, Junzhen Ren, Zhihao Chen, Chang He, Xiaotao Hao, Zhixiang Wei, and Jianhui Hou. 2021.
\newblock Single-junction organic photovoltaic cell with 19\% efficiency.
\newblock \emph{Advanced Materials}, 33(41):2102420.

\bibitem[{Edwards et~al.(2022)Edwards, Lai, Ros, Honke, Cho, and Ji}]{edwards2022translation}
Carl Edwards, Tuan Lai, Kevin Ros, Garrett Honke, Kyunghyun Cho, and Heng Ji. 2022.
\newblock Translation between molecules and natural language.
\newblock \emph{arXiv preprint arXiv:2204.11817}.

\bibitem[{Eibeck et~al.(2021)Eibeck, Nurkowski, Menon, Bai, Wu, Zhou, Mosbach, Akroyd, and Kraft}]{eibeck2021predicting}
Andreas Eibeck, Daniel Nurkowski, Angiras Menon, Jiaru Bai, Jinkui Wu, Li~Zhou, Sebastian Mosbach, Jethro Akroyd, and Markus Kraft. 2021.
\newblock Predicting power conversion efficiency of organic photovoltaics: models and data analysis.
\newblock \emph{ACS omega}, 6(37):23764--23775.

\bibitem[{El~Amine et~al.(2023)El~Amine, Yi, Li, Wang, Xi, and Zhao}]{elamine2023opv20}
Mohamed~Boudia El~Amine, Zhou Yi, Hongying Li, Qiuwang Wang, Jun Xi, and Cunlu Zhao. 2023.
\newblock Latest updates of single-junction organic solar cells up to 20\% efficiency.
\newblock \emph{Energies}, 16:3895.

\bibitem[{Ertl and Schuffenhauer(2009)}]{ertl2009estimation}
Peter Ertl and Ansgar Schuffenhauer. 2009.
\newblock Estimation of synthetic accessibility score of drug-like molecules based on molecular complexity and fragment contributions.
\newblock \emph{Journal of cheminformatics}, 1:1--11.

\bibitem[{Fang et~al.(2022)Fang, Liu, Lei, He, Zhang, Zhou, Wang, Wu, and Wang}]{fang2022geometry}
Xiaomin Fang, Lihang Liu, Jieqiong Lei, Donglong He, Shanzhuo Zhang, Jingbo Zhou, Fan Wang, Hua Wu, and Haifeng Wang. 2022.
\newblock Geometry-enhanced molecular representation learning for property prediction.
\newblock \emph{Nature Machine Intelligence}, 4(2):127--134.

\bibitem[{Fu et~al.(2023)Fu, Fong, Liu, Huang, Lu, Lu, Abdelsamie, Kodalle, Sutter-Fella, Yang, and Li}]{fu2023opv19}
Jiehao Fu, Patrick W.~K. Fong, Heng Liu, Chieh-Szu Huang, Xinhui Lu, Shirong Lu, Maged Abdelsamie, Tim Kodalle, Carolin~M. Sutter-Fella, Yang Yang, and Gang Li. 2023.
\newblock 19.31\% binary organic solar cell and low non-radiative recombination enabled by non-monotonic intermediate state transition.
\newblock \emph{Nature Communications}, 14(1):1760.

\bibitem[{Gillis and Burke(2009)}]{gillis2009iterative}
Eric~P Gillis and Martin~D Burke. 2009.
\newblock Iterative cross-couplng with mida boronates: towards a general platform for small molecule synthesis.
\newblock \emph{Aldrichimica acta}, 42(1):17.

\bibitem[{Greenstein et~al.(2022)Greenstein, Hiener, and Hutchison}]{greenstein2022computational}
Brianna~L Greenstein, Danielle~C Hiener, and Geoffrey~R Hutchison. 2022.
\newblock Computational evolution of high-performing unfused non-fullerene acceptors for organic solar cells.
\newblock \emph{The Journal of Chemical Physics}, 156(17).

\bibitem[{Greenstein and Hutchison(2023)}]{greenstein2023screening}
Brianna~L Greenstein and Geoffrey~R Hutchison. 2023.
\newblock Screening efficient tandem organic solar cells with machine learning and genetic algorithms.
\newblock \emph{The Journal of Physical Chemistry C}, 127(13):6179--6191.

\bibitem[{Guan et~al.(2024)Guan, Li, Xu, Xu, Wang, Xu, Chen, Wang, Zuo, and Chen}]{guan2024opv20}
Shitao Guan, Yaokai Li, Chang Xu, Chenran Xu, Mengting Wang, Yuxi Xu, Qi~Chen, Dawei Wang, Lijian Zuo, and Hongzheng Chen. 2024.
\newblock Self-assembled interlayer enables high-performance organic photovotlaics with power conversion efficiency exceeding 20\%.
\newblock \emph{Advanced Materials}, (2400342):2400342.

\bibitem[{Guo et~al.(2023)Guo, Nan, Liang, Guo, Chawla, Wiest, Zhang et~al.}]{guo2023can}
Taicheng Guo, Bozhao Nan, Zhenwen Liang, Zhichun Guo, Nitesh Chawla, Olaf Wiest, Xiangliang Zhang, et~al. 2023.
\newblock What can large language models do in chemistry? a comprehensive benchmark on eight tasks.
\newblock \emph{Advances in Neural Information Processing Systems}, 36:59662--59688.

\bibitem[{Hachmann et~al.(2011)Hachmann, Olivares-Amaya, Atahan-Evrenk, Amador-Bedolla, S{\'a}nchez-Carrera, Gold-Parker, Vogt, Brockway, and Aspuru-Guzik}]{hachmann2011harvard}
Johannes Hachmann, Roberto Olivares-Amaya, Sule Atahan-Evrenk, Carlos Amador-Bedolla, Roel~S S{\'a}nchez-Carrera, Aryeh Gold-Parker, Leslie Vogt, Anna~M Brockway, and Al{\'a}n Aspuru-Guzik. 2011.
\newblock The harvard clean energy project: large-scale computational screening and design of organic photovoltaics on the world community grid.
\newblock \emph{The Journal of Physical Chemistry Letters}, 2(17):2241--2251.

\bibitem[{Jin et~al.(2023)Jin, Liu, Han, Jiang, Ji, and Han}]{jin2023large}
Bowen Jin, Gang Liu, Chi Han, Meng Jiang, Heng Ji, and Jiawei Han. 2023.
\newblock Large language models on graphs: A comprehensive survey.
\newblock \emph{arXiv preprint arXiv:2312.02783}.

\bibitem[{Kipf and Welling(2016)}]{kipf2016semi}
Thomas~N Kipf and Max Welling. 2016.
\newblock Semi-supervised classification with graph convolutional networks.
\newblock \emph{arXiv preprint arXiv:1609.02907}.

\bibitem[{Kong and Xu(2023)}]{kong2023prediction}
Xiangyu Kong and Bohao Xu. 2023.
\newblock Prediction of photoelectric conversion efficiency of organic photovoltaic materials based on deep learning.
\newblock In \emph{Ninth International Conference on Mechanical Engineering, Materials, and Automation Technology (MMEAT 2023)}, volume 12801, pages 248--253. SPIE.

\bibitem[{Kumar and Chand(2012)}]{kumar2012recent}
Pankaj Kumar and Suresh Chand. 2012.
\newblock Recent progress and future aspects of organic solar cells.
\newblock \emph{Progress in Photovoltaics: Research and applications}, 20(4):377--415.

\bibitem[{Li et~al.(2015)Li, Grillo, and Burke}]{li2015synthesis}
Junqi Li, Anthony~S Grillo, and Martin~D Burke. 2015.
\newblock From synthesis to function via iterative assembly of n-methyliminodiacetic acid boronate building blocks.
\newblock \emph{Accounts of chemical research}, 48(8):2297--2307.

\bibitem[{Liu et~al.(2023{\natexlab{a}})Liu, Ren, and Ren}]{liu2023git}
Pengfei Liu, Yiming Ren, and Zhixiang Ren. 2023{\natexlab{a}}.
\newblock Git-mol: A multi-modal large language model for molecular science with graph.
\newblock \emph{Image, and Text}.

\bibitem[{Liu et~al.(2023{\natexlab{b}})Liu, Nie, Wang, Lu, Qiao, Liu, Tang, Xiao, and Anandkumar}]{liu2023multi}
Shengchao Liu, Weili Nie, Chengpeng Wang, Jiarui Lu, Zhuoran Qiao, Ling Liu, Jian Tang, Chaowei Xiao, and Animashree Anandkumar. 2023{\natexlab{b}}.
\newblock Multi-modal molecule structure--text model for text-based retrieval and editing.
\newblock \emph{Nature Machine Intelligence}, 5(12):1447--1457.

\bibitem[{Liu et~al.(2021)Liu, Wang, Liu, Lasenby, Guo, and Tang}]{liu2021pre}
Shengchao Liu, Hanchen Wang, Weiyang Liu, Joan Lasenby, Hongyu Guo, and Jian Tang. 2021.
\newblock Pre-training molecular graph representation with 3d geometry.
\newblock \emph{arXiv preprint arXiv:2110.07728}.

\bibitem[{Liu et~al.(2023{\natexlab{c}})Liu, Zhang, Xia, Wu, Xie, Qin, Zhang, and Liu}]{liu2023molxpt}
Zequn Liu, Wei Zhang, Yingce Xia, Lijun Wu, Shufang Xie, Tao Qin, Ming Zhang, and Tie-Yan Liu. 2023{\natexlab{c}}.
\newblock Molxpt: Wrapping molecules with text for generative pre-training.
\newblock \emph{arXiv preprint arXiv:2305.10688}.

\bibitem[{Liu et~al.(2023{\natexlab{d}})Liu, Li, Luo, Fei, Cao, Kawaguchi, Wang, and Chua}]{liu2023molca}
Zhiyuan Liu, Sihang Li, Yanchen Luo, Hao Fei, Yixin Cao, Kenji Kawaguchi, Xiang Wang, and Tat-Seng Chua. 2023{\natexlab{d}}.
\newblock Molca: Molecular graph-language modeling with cross-modal projector and uni-modal adapter.
\newblock \emph{arXiv preprint arXiv:2310.12798}.

\bibitem[{Lopez et~al.(2016)Lopez, Pyzer-Knapp, Simm, Lutzow, Li, Seress, Hachmann, and Aspuru-Guzik}]{lopez2016harvard}
Steven~A Lopez, Edward~O Pyzer-Knapp, Gregor~N Simm, Trevor Lutzow, Kewei Li, Laszlo~R Seress, Johannes Hachmann, and Al{\'a}n Aspuru-Guzik. 2016.
\newblock The harvard organic photovoltaic dataset.
\newblock \emph{Scientific data}, 3(1):1--7.

\bibitem[{Lopez et~al.(2017)Lopez, Sanchez-Lengeling, de~Goes~Soares, and Aspuru-Guzik}]{lopez2017design}
Steven~A Lopez, Benjamin Sanchez-Lengeling, Julio de~Goes~Soares, and Al{\'a}n Aspuru-Guzik. 2017.
\newblock Design principles and top non-fullerene acceptor candidates for organic photovoltaics.
\newblock \emph{Joule}, 1(4):857--870.

\bibitem[{McInnes et~al.(1802)McInnes, Healy, and Melville}]{mcinnes1802umap}
Leland McInnes, John Healy, and James Melville. 1802.
\newblock Umap: Uniform manifold approximation and projection for dimension reduction. arxiv 2018.
\newblock \emph{arXiv preprint arXiv:1802.03426}, 10.

\bibitem[{{OpenAI}(2021)}]{gpt3.5}
{OpenAI}. 2021.
\newblock Chatgpt 3.5.
\newblock \url{https://openai.com/gpt-3/}.
\newblock Accessed: January 2022.

\bibitem[{Padula et~al.(2019)Padula, Simpson, and Troisi}]{padula2019combining}
Daniele Padula, Jack~D Simpson, and Alessandro Troisi. 2019.
\newblock Combining electronic and structural features in machine learning models to predict organic solar cells properties.
\newblock \emph{Materials Horizons}, 6(2):343--349.

\bibitem[{Pyzer-Knapp et~al.(2015)Pyzer-Knapp, Li, and Aspuru-Guzik}]{pyzer2015learning}
Edward~O Pyzer-Knapp, Kewei Li, and Alan Aspuru-Guzik. 2015.
\newblock Learning from the harvard clean energy project: The use of neural networks to accelerate materials discovery.
\newblock \emph{Advanced Functional Materials}, 25(41):6495--6502.

\bibitem[{Radford et~al.(2018)Radford, Narasimhan, Salimans, Sutskever et~al.}]{radford2018improving}
Alec Radford, Karthik Narasimhan, Tim Salimans, Ilya Sutskever, et~al. 2018.
\newblock Improving language understanding by generative pre-training.

\bibitem[{Rong et~al.(2020)Rong, Bian, Xu, Xie, Wei, Huang, and Huang}]{rong2020self}
Yu~Rong, Yatao Bian, Tingyang Xu, Weiyang Xie, Ying Wei, Wenbing Huang, and Junzhou Huang. 2020.
\newblock Self-supervised graph transformer on large-scale molecular data.
\newblock \emph{Advances in neural information processing systems}, 33:12559--12571.

\bibitem[{Scharber et~al.(2006)Scharber, M{\"u}hlbacher, Koppe, Denk, Waldauf, Heeger, and Brabec}]{scharber2006design}
Markus~C Scharber, David M{\"u}hlbacher, Markus Koppe, Patrick Denk, Christoph Waldauf, Alan~J Heeger, and Christoph~J Brabec. 2006.
\newblock Design rules for donors in bulk-heterojunction solar cells—towards 10\% energy-conversion efficiency.
\newblock \emph{Advanced materials}, 18(6):789--794.

\bibitem[{Su et~al.(2022)Su, Du, Yang, Zhou, Li, Rao, Sun, Lu, and Wen}]{su2022molecular}
Bing Su, Dazhao Du, Zhao Yang, Yujie Zhou, Jiangmeng Li, Anyi Rao, Hao Sun, Zhiwu Lu, and Ji-Rong Wen. 2022.
\newblock A molecular multimodal foundation model associating molecule graphs with natural language.
\newblock \emph{arXiv preprint arXiv:2209.05481}.

\bibitem[{Sun et~al.(2019)Sun, Li, Li, Wu, Sun, Lu, Xiao, Zhao, and Sun}]{sun2019use}
Wenbo Sun, Meng Li, Yong Li, Zhou Wu, Yuyang Sun, Shirong Lu, Zeyun Xiao, Baomin Zhao, and Kuan Sun. 2019.
\newblock The use of deep learning to fast evaluate organic photovoltaic materials.
\newblock \emph{Advanced Theory and Simulations}, 2(1):1800116.

\bibitem[{Veli{\v{c}}kovi{\'c} et~al.(2017)Veli{\v{c}}kovi{\'c}, Cucurull, Casanova, Romero, Lio, and Bengio}]{velivckovic2017graph}
Petar Veli{\v{c}}kovi{\'c}, Guillem Cucurull, Arantxa Casanova, Adriana Romero, Pietro Lio, and Yoshua Bengio. 2017.
\newblock Graph attention networks.
\newblock \emph{arXiv preprint arXiv:1710.10903}.

\bibitem[{Wang et~al.(2023)Wang, Feng, Dong, Jin, Li, Yuan, and Li}]{wang2023efficient}
Hongshuai Wang, Jie Feng, Zhihao Dong, Lujie Jin, Miaomiao Li, Jianyu Yuan, and Youyong Li. 2023.
\newblock Efficient screening framework for organic solar cells with deep learning and ensemble learning.
\newblock \emph{npj Computational Materials}, 9(1):200.

\bibitem[{Wang et~al.(2022)Wang, Wang, Cao, and Barati~Farimani}]{wang2022molecular}
Yuyang Wang, Jianren Wang, Zhonglin Cao, and Amir Barati~Farimani. 2022.
\newblock Molecular contrastive learning of representations via graph neural networks.
\newblock \emph{Nature Machine Intelligence}, 4(3):279--287.

\bibitem[{Wu et~al.(2020)Wu, Guo, Sun, and Min}]{wu2020machine}
Yao Wu, Jie Guo, Rui Sun, and Jie Min. 2020.
\newblock Machine learning for accelerating the discovery of high-performance donor/acceptor pairs in non-fullerene organic solar cells.
\newblock \emph{npj Computational Materials}, 6(1):120.

\bibitem[{Wu et~al.(2018)Wu, Ramsundar, Feinberg, Gomes, Geniesse, Pappu, Leswing, and Pande}]{wu2018moleculenet}
Zhenqin Wu, Bharath Ramsundar, Evan~N Feinberg, Joseph Gomes, Caleb Geniesse, Aneesh~S Pappu, Karl Leswing, and Vijay Pande. 2018.
\newblock Moleculenet: a benchmark for molecular machine learning.
\newblock \emph{Chemical science}, 9(2):513--530.

\bibitem[{Xiong et~al.(2019)Xiong, Wang, Liu, Zhong, Wan, Li, Li, Luo, Chen, Jiang et~al.}]{xiong2019pushing}
Zhaoping Xiong, Dingyan Wang, Xiaohong Liu, Feisheng Zhong, Xiaozhe Wan, Xutong Li, Zhaojun Li, Xiaomin Luo, Kaixian Chen, Hualiang Jiang, et~al. 2019.
\newblock Pushing the boundaries of molecular representation for drug discovery with the graph attention mechanism.
\newblock \emph{Journal of medicinal chemistry}, 63(16):8749--8760.

\bibitem[{Yang et~al.(2019)Yang, Swanson, Jin, Coley, Eiden, Gao, Guzman-Perez, Hopper, Kelley, Mathea et~al.}]{yang2019analyzing}
Kevin Yang, Kyle Swanson, Wengong Jin, Connor Coley, Philipp Eiden, Hua Gao, Angel Guzman-Perez, Timothy Hopper, Brian Kelley, Miriam Mathea, et~al. 2019.
\newblock Analyzing learned molecular representations for property prediction.
\newblock \emph{Journal of chemical information and modeling}, 59(8):3370--3388.

\bibitem[{Zang et~al.(2023)Zang, Zhao, and Tang}]{zang2023hierarchical}
Xuan Zang, Xianbing Zhao, and Buzhou Tang. 2023.
\newblock Hierarchical molecular graph self-supervised learning for property prediction.
\newblock \emph{Communications Chemistry}, 6(1):34.

\bibitem[{Zeng et~al.(2022)Zeng, Yao, Liu, and Sun}]{zeng2022deep}
Zheni Zeng, Yuan Yao, Zhiyuan Liu, and Maosong Sun. 2022.
\newblock A deep-learning system bridging molecule structure and biomedical text with comprehension comparable to human professionals.
\newblock \emph{Nature communications}, 13(1):862.

\bibitem[{Zhang et~al.(2021)Zhang, Zhu, Zhou, Hao, Qiu, Zhao, Hu, Larson, Zhu, Ma, Tang, Feng, Zhang, Russell, and Liu}]{zhang2021DAmaterials}
Min Zhang, Lei Zhu, Guanqing Zhou, Tianyu Hao, Chaoqun Qiu, Zhe Zhao, Qin Hu, Bryon~W. Larson, Haiming Zhu, Zaifei Ma, Zheng Tang, Wei Feng, Yongming Zhang, Thomas~P. Russell, and Feng Liu. 2021.
\newblock Single-layered organic photovoltaics with double cascading charge transport pathways: 18% efficiencies.
\newblock \emph{Nature Communications}, 12(309).

\bibitem[{Zhao et~al.(2024)Zhao, Liu, Chang, Xu, Fu, Deng, Kong, and Liu}]{zhao2024gimlet}
Haiteng Zhao, Shengchao Liu, Ma~Chang, Hannan Xu, Jie Fu, Zhihong Deng, Lingpeng Kong, and Qi~Liu. 2024.
\newblock Gimlet: A unified graph-text model for instruction-based molecule zero-shot learning.
\newblock \emph{Advances in Neural Information Processing Systems}, 36.

\bibitem[{Zhou et~al.(2023)Zhou, Gao, Ding, Zheng, Xu, Wei, Zhang, and Ke}]{zhou2023uni}
Gengmo Zhou, Zhifeng Gao, Qiankun Ding, Hang Zheng, Hongteng Xu, Zhewei Wei, Linfeng Zhang, and Guolin Ke. 2023.
\newblock Uni-mol: A universal 3d molecular representation learning framework.

\end{thebibliography}

\clearpage

\appendix
\section{Prompt for acquiring text descriptions}
To generate text descriptions for functional modules, we use their SMILES string to query ChatGPT-3.5~\citep{gpt3.5} with this prompt:
\textit{Generate descriptions of  this molecular fragment: [SMILES] focusing on its structural, physical, chemical, and photovoltaic properties. Descriptions should be specific and tailored for organic photovoltaic (OPV) material research. Avoid neutral information.}

\section{Ablation Study}
    \begin{table*}[t]
    \centering
    \small
    \renewcommand{\arraystretch}{1.1}
    \begin{tabular}{p{0.23\textwidth} p{0.07\textwidth} p{0.07\textwidth} p{0.07\textwidth} p{0.07\textwidth} p{0.07\textwidth} p{0.07\textwidth} p{0.07\textwidth} p{0.07\textwidth} p{0.07\textwidth} p{0.07\textwidth} p{0.07\textwidth}}
        \toprule
         & \multicolumn{2}{c}{\textbf{MSE}} & \multicolumn{2}{c}{\textbf{MAE}} & \multicolumn{2}{c}{\textbf{R$^2$}} \\ 
         & \textbf{Avg} & \textbf{Std} & \textbf{Avg} & \textbf{Std} & \textbf{Avg} & \textbf{Std} \\
        \midrule
         
        \textbf{GAT + GAT} & 4.617 & 0.811 & 1.7 & 0.122 & 0.564 & 0.083 \\
        \textbf{GCN + GCN} & 5.924 & 1.08 & 1.94 & 0.19 & 0.407 & 0.128 \\
        \textbf{Attentive FP + GAT} & 4.424 & 0.517 & 1.59 & 0.115 & 0.564 & 0.084 \\
        \textbf{GAT + Attentive FP} & 4.064 & 0.657 & 1.674 & 0.135 & 0.598 & 0.097 \\
        \textbf{Attentive FP + GCN} & 3.481 & 0.922 & 1.457 & 0.181 & 0.648 & 0.091 \\
        \textbf{GCN + Attentive FP} & 5.715 & 0.978 & 1.82 & 0.242 & 0.476 & 0.098 \\
        \textbf{GAT + GCN} & 3.884 & 0.791 & 1.621 & 0.221 & 0.603 & 0.092 \\
        \textbf{GCN + GAT} & 4.672 & 0.98 & 1.734 & 0.133 & 0.559 & 0.105 \\
        \textbf{Attentive FP + Attentive FP} & \textbf{2.327} & 0.801 & \textbf{1.194} & 0.141 & \textbf{0.747} & 0.088 \\
        \bottomrule
        
    \end{tabular}
    \caption{PCE prediction results by \model{} (with textual descriptors being physical and chemical property descriptions) on the collected dataset using various GNN architectures for fragment-level and molecule-level GNNs (leftmost column: fragment-level GNN + molecule-level GNN).}
    \label{table:gnnmodels}
    \end{table*}

    \textbf{GNN architectures for fragment-level GNN and molecule-level GNN.} We experimented with various setups for two GNN models: the fragment-level GNN and the molecule-level GNN. The results are shown in Table~\ref{table:gnnmodels}, indicating that using Attentive FP for both levels gave us the best predictions, scoring an impressive R$^2$ of 0.747 ($\pm$ 0.04).
    
    \textbf{Fusion methods.} For the fusion block, we utilize \textit{average + concat} and \textit{attention + concat} as fusion methods and present the results in Table~\ref{table:concat&attention}.
    Experimental results show that \textit{attention + concat} proves more robust than \textit{average + concat} in 5 out of 6 setups. Hence, for \model{}, we opt for \textit{attention + concat} over \textit{average + concat} as the fusion method for integrating structural descriptors and textual descriptors of functional modules.
    However, \textit{average + concat} outperforms attention when the textual descriptors represent the full description of the functional modules. This is attributed to the lengthier nature of full description text compared to others, resulting in a substantial increase in the number of learnable parameters within the attention block, thereby making it more challenging to learn, particularly in low-resource learning contexts.

    \textbf{Node features for fragment-level GNN.} We also evaluate the effectiveness of utilizing various atomic properties as node features for the fragment-level GNN in \model{}. Experimental results shown in Table~\ref{table:node_attr} demonstrate that solely employing electronegativity as a node feature yields the highest R$^2$ score of 0.747 ($\pm$ 0.04). Consequently, we opt for the electronegativity of atoms as the node attribute for \model{}.
        \begin{table*}[t]
    \centering
    \small
    \renewcommand{\arraystretch}{1.1}
    \begin{tabular}{>{\centering} p{0.06\textwidth} >{\centering} p{0.04\textwidth} >{\centering} p{0.04\textwidth} >{\centering} p{0.04\textwidth} >{\centering} p{0.04\textwidth} >{\centering} p{0.04\textwidth} >{\centering} p{0.04\textwidth} >{\centering} p{0.04\textwidth} >{\centering} p{0.04\textwidth} >{\centering} p{0.04\textwidth} >{\centering} p{0.04\textwidth} >{\centering} p{0.05\textwidth} p{0.05\textwidth}}
        \toprule
        & \multicolumn{2}{c}{\textbf{Full description}} & \multicolumn{2}{c}{\textbf{Structural}} & \multicolumn{2}{c}{\textbf{Physical}} & \multicolumn{2}{c}{\textbf{Chemical}} & \multicolumn{2}{c}{\textbf{Photovoltaic}} & \multicolumn{2}{c}{\textbf{Physical~+~Chemical}} \\
        
        & \textbf{Avg} & \textbf{Att} & \textbf{Avg} & \textbf{Att} & \textbf{Avg} & \textbf{Att} & \textbf{Avg} & \textbf{Att} & \textbf{Avg} & \textbf{Att} & \textbf{Avg} & \textbf{Att} \\
        \midrule
        \textbf{MSE} & 2.437 & 2.878 & 3.346 & 2.66 & 2.762 & 2.561 & 2.541 & 2.308 & 4.274 & 3.317 & 2.343 & 2.327 \\
        \textbf{MAE} & 1.293 & 1.32 & 1.464 & 1.289 & 1.276 & 1.231 & 1.311 & 1.218 & 1.507 & 1.367 & 1.253 & 1.194 \\
        \textbf{R$^2$} & \textbf{0.712} & 0.703 & 0.634 & \textbf{0.725} & 0.712 & \textbf{0.732} & 0.705 & \textbf{0.735} & 0.622 & \textbf{0.659} & 0.719 & \textbf{0.747} \\
        \bottomrule
        
    \end{tabular}
    \caption{\model{}'s PCE prediction results on the collected dataset using \textit{average + concat} (\textbf{Avg}) and \textit{attention + concat} (\textbf{Att}) as fusion operators.}
    \label{table:concat&attention}
    \end{table*}

    \begin{table*}[t]
    \centering
    \small
    \renewcommand{\arraystretch}{1.1}
        \begin{tabular}{p{0.3\textwidth} p{0.06\textwidth} p{0.06\textwidth} p{0.06\textwidth} p{0.06\textwidth} p{0.06\textwidth} p{0.06\textwidth}} 
        \toprule
        \multirow{2}{*}{\textbf{Features}} & \multicolumn{2}{c}{\textbf{MSE}} & \multicolumn{2}{c}{\textbf{MAE}} & \multicolumn{2}{c}{\textbf{R$^2$}} \\
         & \textbf{Avg} & \textbf{Std} & \textbf{Avg} & \textbf{Std} & \textbf{Avg} & \textbf{Std} \\
        \midrule
        Atomic number & 3.343 & 1.03 & 1.353 & 0.159 & 0.679 & 0.107 \\
        Mass & 3.513 & 0.948 & 1.467 & 0.173 & 0.664 & 0.108 \\
        Electronegativity (EN) & \textbf{2.327} & 0.8 & \textbf{1.194} & 0.141 & \textbf{0.747} & 0.088 \\
        EN + hybridization & 2.865 & 0.516 & 1.357 & 0.118 & 0.731 & 0.061 \\
        EN + degree & 3.197 & 0.787 & 1.405 & 0.162 & 0.676 & 0.074 \\
        EN + formal charge & 2.554 & 0.371 & 1.352 & 0.09 & 0.743 & 0.046 \\
        EN + implicit \& explicit valance & 2.951 & 0.532 & 1.337 & 0.129 & 0.7 & 0.069 \\
        EN + is aromatic & 3.199 & 0.803 & 1.366 & 0.171 & 0.668 & 0.084 \\
        All & 2.59 & 0.42 & 1.322 & 0.117 & 0.736 & 0.052 \\
        \bottomrule
        
    \end{tabular}
    \caption{Experimental results of using different kinds of atomic properties as node features for the fragment-level GNN in \model{}, with textual descriptors being physical and chemical property descriptions.}
    \label{table:node_attr}
    \end{table*}

\section{MoleculeNet Tasks}
The details of the four MoleculeNet datasets we utilized for benchmarking \model{} are provided below:
\begin{itemize}
    \item \textbf{BBBP}: Blood-brain barrier penetration (BBBP) assesses the ability of small molecules to cross the blood-brain barrier.
    \item \textbf{BACE}: includes the results of small molecules inhibiting binding to human $\beta$-secretase 1 (BACE-1).
    \item \textbf{ClinTox}: includes information on both the drug's toxicity in clinical trials and its status for FDA approval.
    \item \textbf{SIDER}: The Side Effect Resource (SIDER) includes side effects of drugs affecting 27 organ systems. These drugs encompass not only small molecules but also peptides with molecular weights exceeding 1000.
\end{itemize}

\end{document}